\documentclass{article}

\PassOptionsToPackage{numbers, compress}{natbib}

     \usepackage[final,main]{neurips_2026}

\usepackage[utf8]{inputenc} 
\usepackage[T1]{fontenc}    
\usepackage[colorlinks=true,citecolor=mydarkblue,linkcolor=mydarkblue,urlcolor=mydarkblue]{hyperref}
\usepackage{url}            
\usepackage{booktabs}       
\usepackage{amsfonts}       
\usepackage{nicefrac}       
\usepackage{microtype}      
\usepackage{xcolor}         
\usepackage{pifont}         
\usepackage{tikz}           
\usetikzlibrary{arrows.meta,positioning,fit,calc,decorations.pathreplacing}
\usepackage{multirow}       
\usepackage{enumitem}       
\usepackage{bbm}

\definecolor{mydarkblue}{rgb}{0,0.08,0.45}

\usepackage{wrapfig}
\usepackage{multirow,mathtools}
\usepackage{pifont}
\usepackage{color, colortbl}
\usepackage{bbm}
\usepackage[english]{babel}
\usepackage[autostyle, english=american]{csquotes}
\usepackage{subcaption}
\usepackage{enumitem}
\usepackage{tikz}
\usetikzlibrary{shapes.geometric, arrows.meta, positioning, fit, calc, backgrounds}
\usepackage[capitalize,noabbrev]{cleveref}
\usepackage{algorithm}
\usepackage{algorithmic}
\usepackage{tcolorbox}
\usepackage{soul}

\usepackage{amsmath,amsfonts,bm}

\def\eqref#1{Eq.\,\ref{#1}}

\def\vb{{\bm{b}}}
\def\vc{{\bm{c}}}

\def\vq{{\bm{q}}}

\def\vx{{\bm{x}}}
\def\vy{{\bm{y}}}
\def\vz{{\bm{z}}}

\def\mH{{\bm{H}}}

\def\mP{{\bm{P}}}
\def\mQ{{\bm{Q}}}

\def\mW{{\bm{W}}}

\def\mZ{{\bm{Z}}}

\def\sR{{\mathbb{R}}}

\definecolor{mydarkblue}{rgb}{0,0.08,0.45}
\hypersetup{
  colorlinks=true,
  linkcolor=mydarkblue,
  citecolor=mydarkblue,
  urlcolor=mydarkblue
}

\definecolor{lightgray}{gray}{0.92}
\definecolor{lightblue}{rgb}{0.93,0.95,1.0}

\newcommand{\ours}{\textsc{SpanUQ}}
\newcommand{\bench}{\textsc{SpanUQ-Bench}}

\title{\ours{}: Span-Level Uncertainty Quantification\\for Large Language Model Generation}

\author{%
  Yimeng Zhang$^{1}$\thanks{Corresponding author: \texttt{damonzym@amazon.com}. 
Website: \url{https://damon-demon.github.io/}},\;
  Yingying Zhuang$^{1}$,\;
  Ziyi Wang$^{2}$,\;
  Yuxuan Lu$^{2}$,\;
  Pei Chen$^{1}$, 
  Aman Gupta$^{1}$,\; \\ \bf
  Zhe Su$^{1}$,\;
  Ming Tan$^{1}$,\;
  Zhilin Zhang$^{1}$,\;
  Qun Liu$^{1}$,\; 
  Manikandarajan Ramanathan$^{1}$,  \\ \bf 
  Rajashekar Maragoud$^{1}$,\;
  Edward Vul$^{1}$,\;
  Jing Huang$^{1}$,\;
  Dakuo Wang$^{2}$\\[6pt]
  \normalfont $^{1}$Amazon,\quad $^{2}$Northeastern University
}

\begin{document}

\maketitle

\begin{abstract}
Uncertainty estimation is essential for the trustworthy deployment of large language models (LLMs): it underpins self-refinement and serves as a decision signal for downstream systems, from verifiers to collaborating agents.
However, existing approaches operate at suboptimal granularities: token-level scores lack semantic coherence, while sequence-level scores fail to localize errors.
We formalize \textit{Span-Level Uncertainty Estimation} (SLUE), a new task that targets the natural granularity for uncertainty: semantically coherent text spans, each conveying a single assessable unit of meaning.
To address this task, we introduce \ours{}, a lightweight ($\sim$25M parameter) probe that distills the uncertainty knowledge from expensive multi-sample inference into a single forward pass over LLM hidden states.
\ours{} employs a DETR-style span decoder to simultaneously detect spans and estimate their uncertainty via a Mixture of Beta distribution, trained with a principled combination of Beta NLL regression and contrastive ranking objectives.
We construct \bench{}, the first span-level uncertainty benchmark comprising 20K prompts, $\sim$293K annotated spans, and continuous soft labels derived from multi-sample claim verification.
Experiments on five LLM backbones show that \ours{} consistently achieves the best span-level uncertainty quality (AUROC 0.908–0.944, MAE 0.110–0.129)
, outperforming the strongest probe baseline and all sampling-based methods while being $10$--$20\times$ faster.
Its DETR-based span detector attains 0.910 F1, surpassing the best heuristic by 39.4\%, enabling precise error localization that sequence-level methods cannot provide.
The same architecture ports to five LLMs spanning two model families, with one probe trained per backbone (AUROC 0.908--0.944), and we additionally observe that sequence-level uncertainty is partially decomposable: the learned importance-weighted span composition achieves $\rho_{\text{seq}} = 0.839$, suggesting that span-level estimation subsumes sequence-level as a special case.
The project page is available \href{https://damon-demon.github.io/SpanUQ.html}{damon-demon.github.io/SpanUQ}.
\end{abstract}

\section{Introduction}
\label{sec:intro}

Large language models (LLMs) generate fluent and coherent text across a wide range of tasks, yet their propensity to produce factually incorrect statements, commonly termed hallucinations, remains a critical barrier to deployment in high-stakes domains such as healthcare, legal analysis, and scientific research~\citep{ji2023survey}. A fundamental step toward trustworthy LLM deployment is the ability to estimate \textit{how uncertain} the model is about each piece of information it generates.
Existing approaches to uncertainty estimation operate at two extremes of granularity. \textit{Token-level methods} compute scores for individual tokens using predictive entropy or learned probes~\citep{kadavath2022language, fadeeva2024fact, zhang2025tokur}. While computationally efficient, these scores are semantically incomplete: a single token rarely constitutes a verifiable fact, and function words receive scores that carry no informational value. \textit{Sequence-level methods} produce a single score for the entire response by sampling multiple outputs and measuring their semantic dispersion~\citep{kuhn2023semantic, qiu2024semantic, xiong2023can}. These methods capture meaningful semantic uncertainty but require 
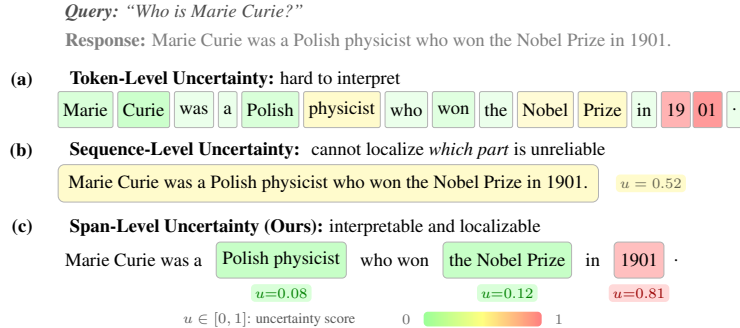
\begin{wrapfigure}{r}{98mm}
\vspace*{-3mm}
\centering
\resizebox{0.7\textwidth}{!}{%
\begin{tikzpicture}[
    every node/.style={font=\small},
    label/.style={font=\small\bfseries, anchor=east},
    score/.style={font=\scriptsize, rounded corners=2pt, inner sep=2pt},
    tok/.style={rounded corners=1pt, inner sep=2.5pt, anchor=west, draw=black!30, line width=0.3pt,
                minimum height=0.61cm},
    span/.style={rounded corners=2pt, inner sep=3pt, anchor=west, line width=0.4pt,
                 minimum height=0.61cm},
]

\def\xstart{0}
\def\rowA{2}    
\def\rowB{0.8}    
\def\rowC{-0.5}   

\node[font=\small\itshape, anchor=west, text=black!70] at (\xstart, 3.5) {%
  \textbf{Query:} ``Who is Marie Curie?''};
\node[font=\small, anchor=west, text=black!50] at (\xstart, 3.0) {%
  \textbf{Response:} Marie Curie was a Polish physicist who won the Nobel Prize in 1901.};

\node[label] at (-0.3, \rowA+0.35) {(a)};
\node[font=\small\bfseries, anchor=west] at (-0.3, \rowA+0.35) {\hspace{1.2em}Token-Level Uncertainty: \textmd{hard to interpret}};

\node[tok, fill=green!15] (t1) at (\xstart, \rowA-0.15) {Marie};
\node[tok, fill=green!20] (t2) at ([xshift=2pt]t1.east) {Curie};
\node[tok, fill=green!8] (t3) at ([xshift=2pt]t2.east) {was};
\node[tok, fill=green!8] (t4) at ([xshift=2pt]t3.east) {a};
\node[tok, fill=green!15] (t5) at ([xshift=2pt]t4.east) {Polish};
\node[tok, fill=yellow!25] (t6) at ([xshift=2pt]t5.east) {physicist};
\node[tok, fill=green!8] (t7) at ([xshift=2pt]t6.east) {who};
\node[tok, fill=green!12] (t8) at ([xshift=2pt]t7.east) {won};
\node[tok, fill=green!8] (t9) at ([xshift=2pt]t8.east) {the};
\node[tok, fill=yellow!20] (t10) at ([xshift=2pt]t9.east) {Nobel};
\node[tok, fill=yellow!25] (t11) at ([xshift=2pt]t10.east) {Prize};
\node[tok, fill=green!8] (t12) at ([xshift=2pt]t11.east) {in};
\node[tok, fill=red!30] (t13) at ([xshift=2pt]t12.east) {19};
\node[tok, fill=red!40] (t14) at ([xshift=1pt]t13.east) {01};
\node[tok, fill=green!8] (t15) at ([xshift=2pt]t14.east) {.};


\node[label] at (-0.3, \rowB+0.35) {(b)};
\node[font=\small\bfseries, anchor=west] at (-0.3, \rowB+0.35) {\hspace{1.2em}Sequence-Level Uncertainty: \textmd{
cannot localize \emph{which part} is unreliable}};

\node[draw=black!30, fill=yellow!25, rounded corners=3pt, inner sep=5pt, anchor=west,
      minimum height=0.45cm] (seq) at (\xstart, \rowB-0.2) {%
  Marie Curie was a Polish physicist who won the Nobel Prize in 1901.};

\node[score, fill=yellow!25, text=yellow!25!black, anchor=west] at ([xshift=8pt]seq.east) {$u = 0.52$};


\node[label] at (-0.3, \rowC+0.35) {(c)};
\node[font=\small\bfseries, anchor=west, text=black] at (-0.3, \rowC+0.35) {\hspace{1.2em}Span-Level Uncertainty (Ours): \textmd{interpretable and localizable}};

\node[anchor=west] (s0) at (\xstart, \rowC-0.2) {Marie Curie was a};

\node[span, fill=green!22, draw=black!30] (s1) at ([xshift=3pt]s0.east) {Polish physicist};
\node[score, fill=green!15, text=green!50!black] at ([yshift=-7pt]s1.south) {$u{=}0.08$};

\node[anchor=west] (s1b) at ([xshift=3pt]s1.east) {who won};

\node[span, fill=green!22, draw=black!30] (s2) at ([xshift=3pt]s1b.east) {the Nobel Prize};
\node[score, fill=green!15, text=green!50!black] at ([yshift=-7pt]s2.south) {$u{=}0.12$};

\node[anchor=west] (s2b) at ([xshift=3pt]s2.east) {in};

\node[span, fill=red!25, draw=black!30, line width=0.5pt] (s3) at ([xshift=3pt]s2b.east) {1901};
\node[score, fill=red!15, text=red!60!black] at ([yshift=-7pt]s3.south) {$u{=}0.81$};

\node[anchor=west] (s3b) at ([xshift=1pt]s3.east) {.};


\def\barY{-1.7}
\node[font=\scriptsize, anchor=west, text=black!65] at (\xstart+2, \barY) {%
    $u \in [0,1]$: uncertainty score};
\shade[left color=green!40, right color=red!50, middle color=yellow!50,
    rounded corners=2pt] (6.2, \barY-0.125) rectangle ++(2, 0.25);
\node[font=\scriptsize, anchor=east, text=black!60] at (6.1, \barY) {$0$};
\node[font=\scriptsize, anchor=west, text=black!60] at (8.3, \barY) {$1$};

\end{tikzpicture}%
}
\caption{Comparison of uncertainty estimation granularities for an LLM response containing a factual error (the Nobel Prize in Physics was awarded in 1903, not 1901). (a)~Token-level methods assign per-token scores that are noisy and hard to interpret. (b)~Sequence-level methods produce a single score, unable to localize which part is unreliable. (c)~Our \emph{span-level} approach identifies claim spans and assigns each an uncertainty score, enabling precise hallucination localization.}
\vspace*{-3mm}
\label{fig:teaser}
\end{wrapfigure}

$10$--$20\times$ inference cost and cannot localize \textit{which parts} of a response are unreliable.
Consider the example in \textbf{Fig.\,\ref{fig:teaser}}: an LLM generates \textit{``Marie Curie was a Polish physicist who won the Nobel Prize in 1901.''} A token-level method~(a) assigns separate scores to every token; function words like ``was'' and ``a'' receive low but meaningless scores, while ``physicist'' receives a moderate score despite being correct, simply because it is a less predictable continuation. A sequence-level method~(b) collapses the entire response into a single score ($u{=}0.52$), averaging correct and incorrect claims into an unactionable middle ground. What users actually need is uncertainty at the level of \textbf{spans}~(c): ``Polish physicist'' ($u{=}0.08$, confident), ``the Nobel Prize'' ($u{=}0.12$, confident), and ``1901'' ($u{=}0.81$, uncertain), since the Nobel Prize in Physics was awarded in 1903, not 1901.
Such localized scores let downstream components (a verifier, a refinement step, or another agent) act on exactly the claims that need attention.
We define a \textbf{span} as a contiguous text segment that conveys a single, coherent unit of meaning whose uncertainty can be independently assessed. In practice, spans often correspond to what the fact-checking literature calls \textit{claims} or \textit{atomic facts}~\citep{min2023factscore}, but the concept is more general: any sub-sentential fragment carrying a self-contained piece of information qualifies. In the example above, ``Polish physicist'',  ``the Nobel Prize'', and ``1901'' are each a span. Unlike tokens, which are sub-word units without standalone meaning, and unlike full sequences, which may bundle multiple assertions of varying reliability, spans are the natural unit for uncertainty estimation: each carries exactly one assessable piece of information, making its uncertainty score directly interpretable and actionable.
This observation leads us to formalize \textbf{Span-Level Uncertainty Estimation (SLUE)} as a new task: given a single LLM forward pass, jointly detect spans and assign continuous uncertainty scores to each.
To address this task, we introduce \ours{} (\textbf{Fig.\,\ref{fig:architecture}}), a lightweight (${\sim}$25M-parameter) probe that \textit{distills} the uncertainty knowledge captured by expensive multi-sample inference into a single forward pass over frozen LLM hidden states.
We construct \bench{}, the first span-level uncertainty benchmark with 20K queries and 293K spans across five domains, and evaluate on five LLMs (Qwen3-14B/8B/4B/30B-A3B, Mistral-7B) spanning 4B--30B parameters with both dense and mixture-of-experts architectures.

Our \underline{\textbf{main contributions}} are as follows.
\textbf{(1)~Problem Formulation.} We formalize SLUE as a new task that bridges the gap between token-level and sequence-level uncertainty estimation, operating at the natural granularity of verifiable claims.
\textbf{(2)~Framework.} We introduce \ours{}, combining a DETR-style span decoder, Mixture of Beta uncertainty estimation, and Uncertainty-Conditioned Iterative Refinement into a lightweight probe (${\sim}$25M parameters) that distills multi-sample uncertainty into a single forward pass.
\textbf{(3)~Benchmark.} We construct \bench{}, the first span-level uncertainty benchmark with continuous soft labels, spanning five domains and 20K queries, together with a multi-sample distillation pipeline that automatically decomposes LLM responses into atomic spans and derives continuous uncertainty labels.
\textbf{(4)~Empirical Results.} Experiments on five LLM backbones show that \ours{} consistently achieves the best span-level uncertainty quality (AUROC 0.908--0.944, MAE 0.110--0.129), outperforming the strongest probe baseline and all sampling-based methods while being $10$--$20\times$ faster. Additionally, we observe that sequence-level uncertainty is partially decomposable into span-level components ($\rho_{\text{seq}} = 0.839$), suggesting that span-level estimation subsumes sequence-level as a special case.

\section{Related Work}
\label{sec:related}

\textbf{Uncertainty Estimation in LLM Generation.}
Uncertainty estimation for LLMs has been studied at multiple granularities.
At the \textit{token level}, predictive entropy serves as a natural baseline~\citep{kadavath2022language}, and lightweight hidden-state probes can improve upon it~\citep{fadeeva2024fact,vazhentsev2025token}.
However, factual correctness is rarely a property of individual tokens: a hallucinated claim such as ``\textit{born in Paris}'' is incorrect as a unit, yet token-level scores assign independent values to each word.
To capture broader semantics, \textit{sequence-level} methods aggregate uncertainty over entire generations.
Semantic Entropy~\citep{kuhn2023semantic} clusters sampled outputs by meaning and computes entropy over the clusters; Kernel Language Entropy~\citep{nikitin2024kernel} replaces discrete clustering with kernel-based similarity.
Other directions include epistemic--aleatoric decomposition~\citep{yadkori2024believe}, claim-level entailment graphs~\citep{da2024llm}, and self-calibration via verbalized confidence~\citep{xiong2023can}.
\textit{Sentence-level} approaches~\citep{chen2024inside} partially localize uncertainty but rely on syntactic boundaries that may not align with semantic units.
UncertaintyRAG~\citep{li2024uncertaintyrag} takes a step toward span-level estimation by computing a signal-to-noise ratio (SNR) over token self-information within fixed sliding windows; however, its spans are defined by rigid windows (length 20, stride 10) rather than learned semantic boundaries, and the resulting scores are used for RAG chunk retrieval rather than generation-quality assessment.
All of these methods produce either a single score per generation or per-token scores, leaving a gap at the \textit{span level}, the natural granularity of factual claims.
\ours{} fills this gap by jointly detecting variable-length spans and assigning 
continuous uncertainty scores to each.

\textbf{Hallucination Detection.}
A parallel line of work asks whether LLM internal representations encode truthfulness signals.
Hidden-state probes can detect hallucinations invisible to output-based methods~\citep{azaria2023internal,orgad2024llms}, and Semantic Entropy Probes~\citep{kossen2024semantic} distill multi-sample signals into single-pass classifiers.
These approaches establish that hidden states are rich sources of factuality information, but they operate at the token or sentence level and produce binary rather than continuous estimates.
From the output side, external-knowledge methods such as FActScore~\citep{min2023factscore} and SAFE~\citep{wei2024long} decompose generations into atomic claims and verify each against a knowledge source, while SelfCheckGPT~\citep{manakul2023selfcheckgpt} measures cross-sample consistency without retrieval.
FAVA~\citep{mishra2024fine} fine-tunes a model to detect and edit hallucinated spans, and the Mu-SHROOM shared task~\citep{vazquez2025semeval} benchmarks span-level hallucination detection, though participating systems rely on heuristic span extraction.
A key distinction is that hallucination detection is binary and often requires external verification, whereas uncertainty estimation produces continuous confidence scores from internal signals alone.
\ours{} bridges the two by learning continuous uncertainty from hidden states, achieving AUROC $>$0.93 for binary hallucination prediction as a byproduct.

\textbf{Span Detection.}
Span detection has a rich history, from named entity recognition~\citep{lample2016neural} to coreference resolution~\citep{lee2017end} and relation extraction~\citep{wadden2019entity}.
Traditional BIO labeling cannot represent overlapping spans, a limitation for uncertainty estimation where multiple claims may coexist within a sentence.
Recent set-prediction approaches inspired by DETR~\citep{carion2020end} address this: CLUE~\citep{wang2024clue} uses a DETR-style architecture for claim-level uncertainty, and HaluNet~\citep{tong2025halunet} applies a similar paradigm to hallucination span detection.
\ours{} adopts the DETR framework but differs in two respects: (1)~we detect \textit{uncertainty-relevant} spans, i.e., any semantically complete unit whose factuality can be assessed, rather than named entities or predefined claim types; and (2)~the span detector is jointly trained with the uncertainty scorer, learning boundaries that are maximally informative for uncertainty estimation.

\vspace{-3mm}
\section{Method: \ours{}}
\label{sec:method}
\vspace{-2mm}

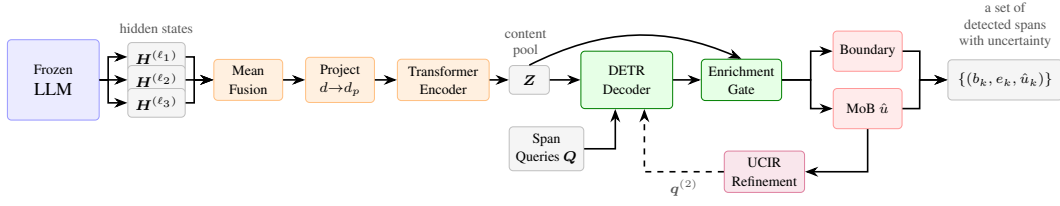
\begin{figure}[t]
\centering
\resizebox{\textwidth}{!}{%
\begin{tikzpicture}[
    >=Stealth,
    box/.style={draw, rounded corners=2pt, minimum height=0.7cm, minimum width=1.6cm, align=center, font=\small},
    llmbox/.style={box, fill=blue!8, draw=blue!40},
    encbox/.style={box, fill=orange!12, draw=orange!50},
    decbox/.style={box, fill=green!10, draw=green!50!black},
    headbox/.style={box, fill=red!8, draw=red!40, minimum width=1.2cm},
    ucirbox/.style={box, fill=purple!10, draw=purple!50},
    databox/.style={draw, rounded corners=2pt, minimum height=0.5cm, align=center, font=\small, fill=gray!8, draw=gray!50},
    arr/.style={->, thick, >=Stealth},
    darr/.style={->, thick, dashed, >=Stealth},
    label/.style={font=\scriptsize, text=gray!70!black},
]

\node[llmbox, minimum width=1.6cm, minimum height=1.4cm] (llm) at (0, 0) {\scriptsize Frozen\\[0pt]\small LLM};

\node[databox, minimum width=1.0cm, minimum height=0.5cm] (l1) at (1.8, 0.4) {\scriptsize $\mH^{(\ell_1)}$};
\node[databox, minimum width=1.0cm, minimum height=0.5cm] (l2) at (1.8, 0) {\scriptsize $\mH^{(\ell_2)}$};
\node[databox, minimum width=1.0cm, minimum height=0.5cm] (l3) at (1.8, -0.4) {\scriptsize $\mH^{(\ell_3)}$};
\node[label] at (1.8, 0.9) {hidden states};

\draw[arr] (llm.east) -- ++(0.15,0) |- (l1.west);
\draw[arr] (llm.east) -- ++(0.15,0) |- (l2.west);
\draw[arr] (llm.east) -- ++(0.15,0) |- (l3.west);

\node[encbox, minimum width=1.2cm] (fuse) at (3.4, 0) {\scriptsize Mean\\[-1pt]\scriptsize Fusion};
\draw[arr] (l1.east) -- ++(0.15,0) |- (fuse.west);
\draw[arr] (l2.east) -- (fuse.west);
\draw[arr] (l3.east) -- ++(0.15,0) |- (fuse.west);

\node[encbox, minimum width=1.2cm] (proj) at (5, 0) {\scriptsize Project\\[-1pt]\scriptsize $d{\to}d_p$};
\draw[arr] (fuse) -- (proj);

\node[encbox, minimum width=1.6cm] (enc) at (6.8, 0) {\scriptsize Transformer\\[-1pt]\scriptsize Encoder};
\draw[arr] (proj) -- (enc);

\node[databox, minimum width=0.7cm, minimum height=0.5cm, inner sep=2pt] (Z) at (8.3, 0) {\scriptsize $\mZ$};
\draw[arr] (enc) -- (Z);

\node[databox, minimum width=1.2cm] (queries) at (8.6, -1.2) {\scriptsize Span\\[-1pt]\scriptsize Queries $\mQ$};

\node[decbox, minimum width=1.6cm, minimum height=1.0cm] (dec) at (10, 0) {\scriptsize DETR\\[-1pt]\scriptsize Decoder};
\draw[arr] (Z) -- (dec);
\draw[arr] (queries.east) -| ([xshift=-0.2cm]dec.south);

\node[decbox, minimum width=1.4cm] (enrich) at (12, 0) {\scriptsize Enrichment\\[-1pt]\scriptsize Gate};
\draw[arr] (dec) -- (enrich);
\draw[arr, bend left=25] (Z.north) to
    node[above, label, xshift=-1.9cm, yshift=-0.5cm] {\shortstack{content\\pool}}
    (enrich.north);

\node[headbox] (bnd) at (14.2, 0.5) {\scriptsize Boundary};
\node[headbox] (mob) at (14.2, -0.5) {\scriptsize MoB $\hat{u}$};
\draw[arr] (enrich.east) -- ++(0.5,0) |- (bnd.west);
\draw[arr] (enrich.east) -- ++(0.5,0) |- (mob.west);

\node[databox, minimum width=2.0cm, minimum height=0.7cm] (out) at (16.6, 0) {\scriptsize $\{(b_k, e_k, \hat{u}_k)\}$};
\node[label, align=center] at (16.6, 0.95) {a set of\\detected spans\\with uncertainty};
\draw[arr] (bnd.east) -- ++(0.3,0) |- (out.west);
\draw[arr] (mob.east) -- ++(0.3,0) |- (out.west);

\node[ucirbox, minimum width=1.4cm] (ucir) at (12.4, -1.6) {\scriptsize UCIR\\[-1pt]\scriptsize Refinement};
\draw[arr] (mob.south) |- (ucir.east);
\draw[darr] (ucir.west) -| node[pos=0.25, below, label] {$\vq^{(2)}$}
    ([xshift=-0.5cm]dec.south east);

\end{tikzpicture}%
}
\caption{Overview of the \ours{} architecture. Given a frozen LLM's hidden states from three selected layers, the probe fuses, projects, and encodes token representations, then uses a DETR-style decoder with learnable span queries to simultaneously detect spans and estimate their uncertainty via Mixture of Beta (MoB) heads. Uncertainty-Conditioned Iterative Refinement (UCIR, dashed) feeds Round-1 estimates back through the shared decoder for a second pass. Each output tuple $(b_k, e_k, \hat{u}_k)$ denotes the beginning position, end position, and estimated uncertainty of a detected span. The trainable probe adds only ${\sim}$25M parameters ($<$0.2\% of the LLM).}
\vspace*{-3mm}
\label{fig:architecture}
\end{figure}

We present \ours{}, a framework that estimates uncertainty at the span level from a single forward pass through a frozen large language model.
The core idea is to \textit{distill} the uncertainty knowledge captured by expensive multi-sample inference into a lightweight module that operates directly on LLM hidden states.
Given a response $\vy = (y_1, \ldots, y_T)$ generated by an LLM for an input prompt $\vx$, \ours{} simultaneously (i)~detects semantically meaningful spans $\{s_k = (b_k, e_k)\}_{k=1}^{K}$ and (ii)~estimates a continuous uncertainty score $u_k \in [0,1]$ for each span, where $0$ denotes full confidence and $1$ denotes maximal uncertainty.
\textbf{Fig.\,\ref{fig:architecture}} provides an overview and \textbf{Fig.\,\ref{fig:span_query_lifecycle}} illustrates the end-to-end span-level uncertainty estimation process with a concrete example (see \textbf{App.\,\ref{app:span_query_lifecycle}} for full design rationale).
Training requires ground-truth spans and uncertainty labels: each span corresponds to a contiguous text segment expressing a single verifiable assertion, and its label $u_k^* \in [0,1]$ measures the fraction of stochastic samples in which the assertion is unsupported (details in \textbf{Sec.\,\ref{sec:exp:data}}).
We first describe the model architecture (\textbf{Sec.\,\ref{sec:method:arch}}), then the training procedure (\textbf{Sec.\,\ref{sec:method:training}}) and inference with iterative refinement (\textbf{Sec.\,\ref{sec:method:inference}}).

\subsection{Model Architecture}
\label{sec:method:arch}

The \ours{} module is a lightweight probe (${\sim}$25M parameters) attached to a frozen LLM backbone.
It consists of five components: multi-layer fusion, a token encoder, a DETR-style span decoder, span feature enrichment, and prediction heads.

\textbf{Multi-Layer Fusion.}
Different LLM layers encode different aspects of uncertainty: early layers capture syntax while mid-to-late layers encode factual knowledge~\citep{orgad2024llms}.
Our layer probing study 
reveals that uncertainty information peaks in layers 20--26 of Qwen3-14B (40 total), following an inverted-U curve.
We select three clustered layers from this peak zone via a lightweight single-layer probing sweep on each backbone; for other models, the optimal zone shifts proportionally with depth (App.\,\ref{app:layer_analysis}, Tab.\,\ref{tab:app_probe_specs}).
Given hidden states $\mH^{(\ell)} \in \sR^{T \times d}$ from selected layers $\mathcal{L} = \{\ell_1, \ell_2, \ell_3\}$, we obtain the fused hidden states $\bar{\mH}$ as follows:
\begin{equation}
  \label{eq:fusion}
  \bar{\mH} = \frac{1}{|\mathcal{L}|} \sum_{\ell \in \mathcal{L}} \mH^{(\ell)} \in \sR^{T \times d}.
\end{equation}

\textbf{Token Encoder.}
The fused hidden states $\bar{\mH}$ are projected to a lower dimension $d_p$ and contextualized through a 2-layer Transformer encoder to obtain the content pool $\mZ$:
\begin{equation}
  \label{eq:encoder}
  \mZ = \text{TransformerEnc}\bigl(\mW_{\text{proj}}\,\bar{\mH} + \mP\bigr) \in \sR^{T \times d_p},
\end{equation}
where $\mW_{\text{proj}} \in \sR^{d_p \times d}$ is a linear projection and $\mP$ denotes sinusoidal positional encodings.

\textbf{DETR-Style Span Decoder.}
Inspired by DETR~\citep{carion2020end}, we formulate span detection as a set prediction problem.
A set of $N{=}32$ learnable \textit{span queries} $\mQ \in \sR^{N \times d_p}$, which are content-agnostic embedding vectors that each learn to bind to one candidate span and analogous to object queries in DETR~\citep{carion2020end}, attend to the token sequence through a 3-layer Transformer decoder.
This capacity comfortably exceeds the typical span count per response (${\sim}$15), ensuring sufficient detection slots while keeping computation tractable.
The decoder produces the refined query set $\hat{\mQ}$:
\begin{equation}
  \label{eq:decoder}
  \hat{\mQ} = \text{TransformerDec}(\mQ, \mZ) \in \sR^{N \times d_p},
\end{equation}
where each layer applies self-attention among span queries (modeling inter-span dependencies) followed by cross-attention over $\mZ$.
This design naturally handles overlapping spans and enables parallel detection in a single pass.
Following DETR, we use the Hungarian algorithm~\citep{kuhn1955hungarian} to find an optimal one-to-one assignment $\hat{\sigma}$ between predicted and ground-truth spans.
The matching cost for each prediction--target pair sums three terms: (i)~L1 boundary distance plus Generalized IoU~\citep{rezatofighi2019generalized}, (ii)~absolute uncertainty error, and (iii)~binary cross-entropy of the validity score (details in \textbf{App.\,\ref{app:loss}}).

\textbf{Span Feature Enrichment.}
The decoder queries capture \textit{where} a span is but lack direct access to the \textit{content} within the predicted region.
To bridge this gap, we introduce a differentiable soft boundary mask that extracts span-level content from the token features $\mZ$.
Given predicted boundaries $(\hat{b}_k, \hat{e}_k)$, we define a mask $m_k(t)$ over token positions:
\begin{equation}
  \label{eq:soft_mask}
  m_k(t) = \sigma\!\bigl(\tau(t - \hat{b}_k)\bigr) \cdot \sigma\!\bigl(\tau(\hat{e}_k - t)\bigr),
\end{equation}
where $\sigma$ is the sigmoid function and $\tau$ controls boundary sharpness.
Intuitively, $m_k(t) \approx 1$ for tokens inside the predicted span and decays smoothly to $0$ outside, enabling gradient flow through the boundary predictions.
We then pool token features via mask-weighted attention, yielding content vector $\vc_k$:
\begin{equation}
  \label{eq:attn_pool}
  \vc_k = \sum_{t=1}^{T} \alpha_{k,t} \cdot \vz_t, \quad
  \alpha_{k,t} \propto m_k(t) \cdot \exp\!\bigl(\hat{\vq}_k^\top \mW_a \vz_t / \sqrt{d_p}\bigr),
\end{equation}
where the mask $m_k(t)$ restricts attention to the predicted region while the query--token dot product selects the most informative tokens within it.
The pooled content is fused with the query via a gated residual connection to produce the enriched query $\tilde{\vq}_k$:
\begin{equation}
  \label{eq:gate_fusion}
  \tilde{\vq}_k = \hat{\vq}_k + \sigma\!\bigl(\mW_g \hat{\vq}_k + \vb_g\bigr) \odot \vc_k,
\end{equation}
which adaptively controls how much span content contributes. This gating is especially useful for short spans where the query alone may suffice.

\textbf{Prediction Heads.}
Each enriched query $\tilde{\vq}_k$ is passed to three heads.
\textit{Span regression} predicts normalized boundaries $(\hat{b}_k, \hat{e}_k) = \sigma(\text{MLP}_{\text{reg}}(\hat{\vq}_k))$ from the pre-enrichment query to avoid circular dependency.
\textit{Validity classification} predicts $p_k^{\text{valid}} = \sigma(\text{MLP}_{\text{val}}(\tilde{\vq}_k))$ to distinguish real spans from unused slots (i.e., queries not matched to any span).
\textit{Mixture of Beta (MoB) uncertainty estimation} models span uncertainty as:
\begin{equation}
  \label{eq:mob}
  p(u_k \mid \tilde{\vq}_k) = \sum_{j=1}^{J} \pi_{k,j} \cdot \text{Beta}(u_k;\, \alpha_{k,j},\, \beta_{k,j}),
\end{equation}
with $J = 3$ components and $\alpha_{\min} = 0.5$ to allow U-shaped distributions that capture the bimodal nature of uncertainty.
The predicted score is the mixture mean $\hat{u}_k = \sum_j \pi_{k,j} \alpha_{k,j} / (\alpha_{k,j} + \beta_{k,j})$.
Span-level scores are aggregated to a sequence-level estimate via importance-weighted pooling:
\begin{equation}
  \label{eq:seq_agg}
  \hat{u}_{\text{seq}} = \sum_{k=1}^{N} w_k \cdot \hat{u}_k, \quad
  w_k = \text{softmax}_k\bigl(p_k^{\text{valid}} \cdot \text{MLP}_{\text{imp}}(\tilde{\vq}_k)\bigr),
\end{equation}
providing a principled decomposition of sequence-level uncertainty into span-level components.

\subsection{Training}
\label{sec:method:training}

\textbf{Two-Phase Strategy.}
We adopt a two-phase training strategy:
(1)~a \textit{warmup} phase (epochs 1--$E_w$) that trains only span detection components (regression + validity) to establish reliable boundaries; and
(2)~a \textit{joint} phase (epochs $E_w$+1--$E$) that activates all components with the full loss:
\begin{equation}
  \label{eq:total_loss}
  \mathcal{L} = \underbrace{\lambda_{\text{reg}} \mathcal{L}_{\text{reg}} + \lambda_{\text{val}} \mathcal{L}_{\text{val}}}_{\text{span detection}} + \underbrace{\lambda_{\text{uq}} \mathcal{L}_{\text{MoB}} + \lambda_{\text{con}} \mathcal{L}_{\text{con}} + \lambda_{\text{rank}} \mathcal{L}_{\text{rank}}}_{\text{uncertainty estimation}} + \lambda_{\text{aux}} \mathcal{L}_{\text{aux}}.
\end{equation}

\textbf{Loss Components.}
The \textit{span regression loss} $\mathcal{L}_{\text{reg}}$ combines L1 and Generalized IoU losses over Hungarian-matched span pairs.
The \textit{validity loss} $\mathcal{L}_{\text{val}}$ is binary cross-entropy over all span queries.
The \textit{MoB NLL loss} $\mathcal{L}_{\text{MoB}}$ maximizes the likelihood of ground-truth uncertainty under the predicted mixture (Eq.\,\ref{eq:mob}):
\begin{equation}
  \label{eq:mob_nll}
  \mathcal{L}_{\text{MoB}} = -\frac{1}{M} \sum_{i=1}^{M} \log \sum_{j=1}^{J} \pi_{i,j} \cdot \text{Beta}(u_{\hat{\sigma}(i)}^*;\, \alpha_{i,j},\, \beta_{i,j}),
\end{equation}
where $M$ is the number of matched spans and $\hat{\sigma}$ the Hungarian assignment.
The \textit{contrastive ranking loss} $\mathcal{L}_{\text{rank}}$ preserves ordinal structure among span pairs within the same sequence:
\begin{equation}
  \label{eq:rank}
  \mathcal{L}_{\text{rank}} = \frac{1}{|\mathcal{P}|} \sum_{(i,j) \in \mathcal{P}} \max\bigl(0,\, -\text{sign}(u_i^* - u_j^*) \cdot (\hat{u}_i - \hat{u}_j) + m\bigr),
\end{equation}
where $\mathcal{P}$ is a set of span pairs with sufficiently different ground-truth uncertainty (see \textbf{App.\,\ref{app:loss}} for the stratified sampling strategy) and $m$ is the margin.
The \textit{consistency loss} $\mathcal{L}_{\text{con}} = \lVert \hat{u}_{\text{seq}} - u_{\text{seq}}^* \rVert_2^2$ enforces agreement between span-aggregated and ground-truth sequence-level uncertainty (with detached gradients to prevent interference).
\textit{Auxiliary losses} $\mathcal{L}_{\text{aux}}$ apply the same objectives to intermediate decoder layers following DETR.
Full formulations of all components and loss weights are in \textbf{App.\,\ref{app:loss}}.

\subsection{Inference with Iterative Refinement}
\label{sec:method:inference}

At inference time, \ours{} requires only a single LLM forward pass to extract hidden states, followed by a lightweight forward pass through the \ours{} module.
No sampling, multiple forward passes, or external knowledge retrieval is needed.
Since initial uncertainty estimates from a single decoder pass may be coarse,
\textbf{Uncertainty-Conditioned Iterative Refinement (UCIR)} feeds Round~1 estimates back into the decoder for a refinement round:
\begin{equation}
  \label{eq:ucir}
  \vq_k^{(2)} = \vq_k^{(1)} + \text{MLP}_{\text{cond}}\bigl(\hat{u}_k^{(1)},\, \hat{\nu}_k^{(1)},\, p_k^{(1)}\bigr),
\end{equation}
where $\hat{\nu}_k^{(1)}$ is the effective precision and $p_k^{(1)}$ the validity probability.
The conditioned queries pass through the \textit{same} decoder (shared weights), giving the final uncertainty estimate $\hat{u}_k$:
\begin{equation}
  \label{eq:ucir_residual}
  \hat{u}_k = \alpha \cdot \hat{u}_k^{(2)} + (1 - \alpha) \cdot \hat{u}_k^{(1)},
\end{equation}
with $\alpha = 0.7$.
Both rounds contribute to the training loss (Round~1 with reduced weight).
UCIR adds $<$15\% overhead relative to \ours{}'s own base decoder (not to the LLM forward pass) since the decoder is shared.
%
The \textbf{full inference pipeline}: 
1)~extract hidden states from layers $\mathcal{L}$; 
2)~fuse, project, and encode; 
3)~decode span queries and predict boundaries, validity, and uncertainty; 
4)~apply UCIR; 
5)~filter by validity threshold and output $\{(b_k, e_k, \hat{u}_k)\}$.
The entire module adds $<$3\% latency to the LLM forward pass.

\section{Experiments}
\label{sec:exp}
\vspace*{-3mm}

We evaluate \ours{} on span-level uncertainty estimation quality, calibration, span detection accuracy, and ablation of each architectural component.
All main experiments use Qwen3-14B as the backbone LLM; the full model adds only 24.4M trainable parameters ($<$0.2\% of the LLM).

\subsection{Dataset: \bench{}}
\label{sec:exp:data}

Evaluating span-level uncertainty requires ground-truth annotations at the span granularity, which no existing benchmark provides.
We construct \bench{}, a benchmark with labels derived from multi-sample distillation and validated by an independent human annotation study.

\textbf{Data Collection.}
We collect LLM-generated texts from five domains using Qwen3-14B as the primary backbone:
1)~\textit{Long-form QA} (42\%) from Natural Questions~\citep{kwiatkowski2019natural};
2)~\textit{TriviaQA} (25\%) from the unfiltered split~\citep{joshi2017triviaqa};
3)~\textit{ELI5} (15\%) from Reddit explanatory questions;
4)~\textit{Biography generation} (10\%) following the FActScore protocol~\citep{min2023factscore}, with entities stratified by Wikipedia popularity; and
5)~\textit{FELM} (8\%) aggregating challenging examples from FELM~\citep{zhao2023felm}, TruthfulQA~\citep{lin2022truthfulqa}, and HaluEval~\citep{li2023halueval}.
The resulting dataset contains 
20k
prompts spanning factoid, explanatory, and reasoning questions.

\textbf{Labels via Multi-Sample Distillation.}
\label{sec:method:training:data}
For each prompt, we generate a \textit{primary response} with greedy decoding ($T = 0$), which serves as the probe input (providing hidden states and span positions).
Separately, we sample $S = 20$ diverse responses at temperature $T = 1.0$.
We decompose \emph{the primary response} into atomic claims with character-level span offsets using an LLM judge, then ask the judge, for every claim and every one of the $S$ samples, whether that sample \textsc{supports}, \textsc{contradicts}, or \textsc{does not mention} the claim.
The soft label $u_k^* \in [0,1]$ is the contradiction rate among the samples that take a position on span $s_k$:
\begin{equation}
  \label{eq:empirical_unc}
  u_k^* = \frac{n_k^{\text{con}}}{n_k^{\text{sup}} + n_k^{\text{con}}},
  \qquad n_k^{\text{sup}} + n_k^{\text{con}} > 0,
\end{equation}
where $n_k^{\text{sup}}$ and $n_k^{\text{con}}$ count samples that support and contradict $s_k$.
Samples that do not mention the claim are excluded rather than counted as evidence, so the pipeline abstains on silence, and spans no sample addresses are dropped (\textbf{App.\,\ref{app:distillation}}).
Crucially, \textbf{no retrieval or external knowledge source participates in label construction}: $u_k^*$ is a pure cross-sample consistency statistic.
The probe thus learns to \textit{distill} the uncertainty knowledge from 20 stochastic samples into a single deterministic forward pass.
For Qwen3-14B, the resulting training set contains ${\sim}$17{,}500 examples with ${\sim}$256K valid span-level soft labels (training split only; the full filtered benchmark contains ${\sim}$293K spans across all splits, see \textbf{App.\,\ref{app:benchmark_stats}}).
The distribution is bimodal: most spans cluster near $u \approx 0$ (consistently correct) or $u \approx 1$ (consistently wrong), with a meaningful tail of intermediate values.
%
For evaluation, we apply the same distillation pipeline to responses generated with greedy decoding ($T = 0$).

\textbf{Label Validity.}
Five independent annotators judged 2{,}000 randomly sampled spans as factually \textsc{correct} or \textsc{incorrect}.
Binarizing $u^*$ at $0.5$ gives Cohen's $\kappa = 0.618$ and 89.4\% raw agreement, and human-judged error rises monotonically across $u^*$ bins (5.8\% $\to$ 87.6\%; $\rho = 0.714$); these human labels are the ground truth for \textbf{Sec.\,\ref{sec:exp:risk_coverage}}.
Protocol, per-bin results, and a root-cause audit of the disagreements are in \textbf{App.\,\ref{app:human_study}} and \textbf{\ref{app:fp_audit}}.
Data collection, the distillation pipeline, test set construction, and benchmark statistics are detailed in \textbf{App.\,\ref{app:data_collection}}--\textbf{\ref{app:benchmark_stats}}.

\subsection{Experimental Setup}
\label{sec:exp:setup}

\textbf{Models.}
All core experiments use \textbf{Qwen3-14B} (dense, 14B parameters)~\citep{yang2025qwen3} as the primary backbone.
Hidden states are extracted from layers 22, 24, and 26 (out of 40 total) and fused via element-wise mean, following our layer importance probing study (\textbf{App.\,\ref{app:layer_analysis}}).
The LLM is frozen throughout; only the \ours{} module is trained.
To evaluate \emph{architectural portability}, we train a separate \ours{} probe per backbone on four additional LLMs: Qwen3-8B, Qwen3-4B, Qwen3-30B-A3B (MoE), and Mistral-7B-Instruct-v0.3.
We do not claim zero-shot cross-model transfer: each probe is trained on its own backbone's hidden states, analogous to linear probing, since backbones differ in hidden dimension and in which layers carry the signal.
Each model generates its own responses and hidden states; the data construction pipeline (\textbf{Sec.\,\ref{sec:exp:data}}) and training procedure are identical, with layer selection and projection dimensions tuned per model via hyperparameter sweep (\textbf{App.\,\ref{app:layer_analysis}}).
Full model specifications are in \textbf{App.\,\ref{app:model_specs}}.

\textbf{Training Details.}
We train on 17{,}494 samples (${\sim}$256K valid spans, training split) with batch size 16 on a single NVIDIA H100 80GB GPU.
Training proceeds in two phases: a 15-epoch \emph{warmup} phase that trains only the span detection components (token encoder, DETR decoder, regression head, validity head), followed by a 25-epoch \emph{joint} phase that also activates the MoB uncertainty head, contrastive ranking loss, and UCIR refinement.
We use AdamW with learning rate $10^{-4}$, weight decay $0.01$, and cosine annealing.
Early stopping monitors dev AUROC with patience 5.
Training curves and computational cost analysis are in \textbf{App.\,\ref{app:training_curves}} and \textbf{\ref{app:cost}}.

\textbf{Evaluation Metrics.}
We report metrics spanning discrimination, regression, and correlation.
\textbf{AUROC} measures binary hallucination detection (threshold $u \geq 0.5$).
\textbf{MAE} is the mean absolute error between predicted $\hat{u}$ and ground-truth $u^*$.
\textbf{Spearman~($\rho$)} correlation captures ranking quality at both span and sequence levels; we abbreviate $\rho_s \equiv \rho_{\text{span}}$ in tables.
$\textbf{AUROC}_{@0.3}$ is the sequence-level AUROC treating sequences with mean uncertainty ${\geq}\,0.3$ as positive.
\textbf{Detection F1} is the harmonic mean of span detection precision (\textbf{Prec}) and recall (\textbf{Rec}) under IoU $\geq 0.3$ matching; this threshold applies to detection \emph{only} --- the uncertainty metrics use Hungarian-matched pairs with no IoU cutoff.
\textbf{ECE} (Expected Calibration Error) measures the gap between predicted uncertainty and observed error rate.

\textbf{Baselines.}
We compare against methods spanning four granularity levels.
\textit{Token-level}: \emph{Token Entropy} (mean token-level entropy from output logits, aggregated over GT spans) and \emph{MLP Probe} (3-layer MLP trained on the same span features as \ours{}, with oracle span boundaries).
\textit{Sequence-level}: \emph{Semantic Entropy} (SE)~\citep{kuhn2023semantic} (NLI-based clustering of 10 sampled responses), \emph{SelfCheckGPT-NLI}~\citep{manakul2023selfcheckgpt} (NLI-based self-consistency over 10 samples), \emph{Verbalized Confidence} (prompting the LLM to self-assess factual confidence on a 0--100 scale), \emph{P(True)}~\citep{kadavath2022language} (probability the LLM assigns to ``True'' when asked whether its answer is correct), \emph{INSIDE}~\citep{chen2024inside} (eigenvalue decomposition of Jaccard similarity matrices over sampled responses), and \emph{KLE}~\citep{nikitin2024kernel} (kernel-based language entropy using RBF kernels on NLI embeddings).
\textit{Claim-level}: \emph{FActScore}~\citep{min2023factscore} (fraction of claims not supported by self-verification, computed from our claim decomposition pipeline).
For end-to-end evaluation (\textbf{Sec.\,\ref{sec:exp:span_detection}}), we also pair the MLP probe with heuristic span detectors including sliding windows, sentence boundaries, NER entities, and token-level thresholding.
For span-level evaluation (\textbf{Tab.\,\ref{tab:span}}), sequence- and claim-level baselines broadcast their scores to all spans within the sequence.
For sequence-level evaluation (\textbf{Tab.\,\ref{tab:seq}} in the Appendix), \ours{} aggregates span predictions via importance-weighted top-$k$.

\subsection{Main Results}
\label{sec:exp:main}

\begin{table*}[t]
  \caption{Span-level uncertainty estimation across five LLM backbones spanning two model families, three scales, and both dense and MoE architectures.
    Token-level methods aggregate token scores within each span;
    sequence- and claim-level methods broadcast a single score to all spans in the sequence.
    $^{*}$MLP Probe uses ground-truth span boundaries (oracle setting). Best in \textbf{bold}, second best \underline{underlined} (per model).
    }
  \label{tab:span}
  \centering
  \resizebox{\textwidth}{!}{%
  \begin{tabular}{l ccc ccc ccc ccc ccc}
    \toprule
    & \multicolumn{3}{c}{\textbf{Qwen3-14B}} & \multicolumn{3}{c}{\textbf{Qwen3-8B}} & \multicolumn{3}{c}{\textbf{Qwen3-4B}} & \multicolumn{3}{c}{\textbf{Qwen3-30B-A3B}} & \multicolumn{3}{c}{\textbf{Mistral-7B}} \\
    \cmidrule(lr){2-4} \cmidrule(lr){5-7} \cmidrule(lr){8-10} \cmidrule(lr){11-13} \cmidrule(lr){14-16}
    \textbf{Method} & AUROC$\uparrow$ & MAE$\downarrow$ & $\rho_s\!\uparrow$ & AUROC$\uparrow$ & MAE$\downarrow$ & $\rho_s\!\uparrow$ & AUROC$\uparrow$ & MAE$\downarrow$ & $\rho_s\!\uparrow$ & AUROC$\uparrow$ & MAE$\downarrow$ & $\rho_s\!\uparrow$ & AUROC$\uparrow$ & MAE$\downarrow$ & $\rho_s\!\uparrow$ \\
    \midrule
    Token Entropy & 0.603 & 0.230 & 0.097 & 0.580 & 0.226 & 0.111 & 0.575 & 0.272 & 0.119 & 0.592 & 0.196 & 0.097 & 0.569 & 0.231 & 0.109 \\
    MLP Probe$^{*}$ & \underline{0.881} & \underline{0.139} & \underline{0.575} & \underline{0.884} & \underline{0.149} & \underline{0.612} & \underline{0.873} & \underline{0.179} & \underline{0.626} & \underline{0.893} & \underline{0.120} & \underline{0.575} & \underline{0.863} & \underline{0.146} & \underline{0.562} \\
    Verbalized Conf. & 0.593 & 0.187 & 0.131 & 0.574 & 0.217 & 0.155 & 0.595 & 0.267 & 0.141 & 0.598 & 0.167 & 0.117 & 0.498 & 0.171 & \llap{$-$}0.026 \\
    SelfCheckGPT-NLI & 0.808 & 0.158 & 0.471 & 0.801 & 0.182 & 0.473 & 0.782 & 0.215 & 0.479 & 0.813 & 0.147 & 0.455 & 0.816 & 0.154 & 0.499 \\
    P(True) & 0.526 & 0.197 & 0.087 & 0.546 & 0.233 & 0.121 & 0.529 & 0.273 & 0.122 & 0.541 & 0.194 & 0.123 & 0.505 & 0.199 & 0.005 \\
    FActScore & 0.625 & 0.178 & 0.181 & 0.762 & 0.197 & 0.405 & 0.714 & 0.233 & 0.390 & 0.701 & 0.169 & 0.306 & 0.661 & 0.182 & 0.297 \\
    HaloScope~\citep{du2024haloscope} & 0.834 & 0.156 & 0.501 & 0.828 & 0.180 & 0.511 & 0.825 & 0.208 & 0.539 & 0.835 & 0.145 & 0.485 & 0.776 & 0.176 & 0.426 \\
    HaMI~\citep{niu2026robust} & 0.775 & 0.187 & 0.445 & 0.814 & 0.185 & 0.483 & 0.821 & 0.210 & 0.530 & 0.798 & 0.151 & 0.422 & 0.764 & 0.178 & 0.407 \\
    \midrule
    \ours{} & \textbf{0.939} & \textbf{0.110} & \textbf{0.685} & \textbf{0.930} & \textbf{0.129} & \textbf{0.692} & \textbf{0.944} & \textbf{0.126} & \textbf{0.754} & \textbf{0.936} & \textbf{0.110} & \textbf{0.647} & \textbf{0.908} & \textbf{0.126} & \textbf{0.637} \\
    \bottomrule
  \end{tabular}}%
\end{table*}

\textbf{Span-Level Analysis (Tab.\,\ref{tab:span}).}
\ours{} achieves an AUROC of \textbf{0.939}, outperforming the strongest probe baseline (MLP Probe, 0.881) by 5.8 absolute points and the best multi-sample method (SelfCheckGPT-NLI, 0.808) by 13.1 points, while requiring only a single forward pass.
Sequence-level methods suffer from a fundamental granularity mismatch when broadcast to spans: SelfCheckGPT-NLI achieves reasonable AUROC (0.808) but assigns the same score to all spans within a sequence, limiting its span-level Spearman correlation to 0.471.
FActScore (0.625 AUROC) achieves only 0.181 Spearman correlation, indicating that its binary self-verification outputs fail to capture fine-grained uncertainty ranking.
P(True) performs near chance (AUROC = 0.526), because the LLM overwhelmingly predicts ``True'' (93.8\% of claims), yielding almost no discriminative signal.
\ours{} also achieves the lowest MAE (\textbf{0.110}), owing to the Mixture of Beta output distribution (\textbf{App.\,\ref{app:mob_analysis}}).
The hidden-state detectors HaloScope and HaMI are the closest prior work, but both are sequence-level and must broadcast one score to every span: their AUROC (0.820 and 0.794 on average) rivals multi-sample methods at no extra generation cost, yet their span-level ranking is capped ($\rho_s$ 0.492 and 0.457 vs.\ 0.683 for \ours{}), and \ours{} leads HaloScope by 11.1 AUROC points on average (\textbf{App.\,\ref{app:new_baselines}}).
These trends hold consistently across all five backbones, including three Qwen3 scales (14B/8B/4B) and entity popularity strata (\textbf{App.\,\ref{app:entity_popularity}}).
The Spearman correlation of \textbf{0.685} represents a 19.1\% relative improvement over the MLP Probe (0.575), demonstrating that \ours{} preserves fine-grained uncertainty ordering across spans.

\textbf{Sequence-Level Aggregation.}
When aggregated to the sequence level (\textbf{Tab.\,\ref{tab:seq}} in the Appendix), \ours{} achieves $\text{AUROC}_{@0.3} = 0.948$ and Spearman $\rho = 0.851$, competitive with or surpassing all dedicated sequence-level methods including SelfCheckGPT-NLI ($\rho = 0.805$) which requires $10{\times}$ generation cost.
The MLP Probe with oracle span boundaries slightly outperforms \ours{} at the sequence level ($\rho = 0.858$), but this advantage disappears when ground-truth spans are unavailable. \ours{}'s span-level superiority (AUROC: 0.939 vs.\ 0.881) and end-to-end detection capability (\textbf{Sec.\,\ref{sec:exp:span_detection}}) make it the only practical solution for fine-grained uncertainty estimation.
Notably, this span-to-sequence decomposability is consistent across all five backbones (Spearman $\rho_{\text{seq}} \geq 0.775$; \textbf{App.\,\ref{app:decomposability}}), confirming that fine-grained span predictions capture sufficient structure for reliable sequence-level inference.
More details can be found in \textbf{App.\,\ref{app:seq_level}}.
Qualitative examples illustrating span-level predictions are in \textbf{App.\,\ref{app:qualitative}}.

\vspace*{-3mm}
\subsection{Human-Grounded Risk--Coverage Analysis}
\label{sec:exp:risk_coverage}
\vspace*{-3mm}

AUROC measures ranking quality against pipeline labels, but deployment asks a different question: if we reject the most uncertain spans, how much factual error do we actually remove?
We answer it with a risk--coverage analysis whose \textbf{ground truth is the independent human annotation} (Sec.\,\ref{sec:exp:data}), not the pipeline label the probe was trained on: risk is the human-judged error rate among retained spans, coverage the fraction retained, AURC the area under the curve (lower is better), and PRR the fraction of the random-to-oracle gap a method closes.

\begin{table}[t]
  \caption{Risk--coverage on human-annotated factual labels (Qwen3-14B, seed 42). Risk is the human-judged error rate among retained spans. 
  $^{*}$MLP Probe uses ground-truth span boundaries (oracle setting) and is therefore not deployable. Best in \textbf{bold}.
  }
  \label{tab:risk_coverage}
  \centering
  \scriptsize
  \begin{tabular}{l cccc cc}
    \toprule
    \textbf{Method} & \textbf{Risk@100\%} & \textbf{Risk@80\%} & \textbf{Risk@60\%} & \textbf{Risk@40\%} & \textbf{AURC}$\downarrow$ & \textbf{PRR}$\uparrow$ \\
    \midrule
    \ours{} (ours)     & 18.6\% & \textbf{10.5\%} & \textbf{7.9\%} & \textbf{6.3\%} & \textbf{0.076} & \textbf{0.587} \\
    MLP Probe$^{*}$    & 18.6\% & 12.8\% & 9.4\% & 7.1\% & 0.098 & 0.470 \\
    Token Entropy      & 18.6\% & 15.2\% & 13.1\% & 11.8\% & 0.138 & 0.254 \\
    FActScore          & 18.6\% & 15.8\% & 13.9\% & 12.4\% & 0.145 & 0.216 \\
    SelfCheckGPT-NLI   & 18.6\% & 16.1\% & 14.5\% & 12.9\% & 0.151 & 0.184 \\
    Verbalized Conf.   & 18.6\% & 16.4\% & 15.0\% & 13.5\% & 0.155 & 0.162 \\
    P(True)            & 18.6\% & 16.8\% & 15.3\% & 14.0\% & 0.159 & 0.140 \\
    \midrule
    Random             & 18.6\% & 18.6\% & 18.6\% & 18.6\% & 0.185 & --- \\
    \bottomrule
  \end{tabular}
  \vspace{-6mm}
\end{table}

\textbf{Selective Prediction (Tab.\,\ref{tab:risk_coverage}).}
Rejecting only the top-20\% most uncertain spans cuts the factual error rate on retained spans from 18.6\% to \textbf{10.5\%}, a 43.9\% relative reduction, and \ours{} attains the best AURC (0.076 vs.\ 0.185 random) and PRR (0.587) of all methods.
An operating point can be read straight off the curve: a 10\% error budget corresponds to ${\sim}$80\% coverage (\textbf{App.\,\ref{app:thresholds}}).
Because this risk reduction is measured against \emph{human} judgments that played no role in training, it also shows the probe is not merely reproducing pipeline noise --- a probe that had memorized pipeline artifacts would not yield a monotone decrease in human-judged error as its own score rises (\textbf{App.\,\ref{app:not_imitation}}).

\vspace*{-3mm}
\subsection{End-to-End Span Detection}
\label{sec:exp:span_detection}
\vspace*{-3mm}

A key advantage of \ours{} is its ability to \emph{simultaneously} detect uncertain spans and estimate their uncertainty, without requiring pre-defined span boundaries.
\textbf{Tab.\,\ref{tab:span_detection}} compares \ours{}'s DETR-based span detector against heuristic alternatives, where each method's detected spans are scored by the same MLP probe for a controlled comparison.

\begin{wraptable}{r}{86mm}
\vspace{-4mm}
  \centering
  \caption{End-to-end span detection comparison. Each method detects spans, scored by an identical MLP probe. \textbf{Bold}: best; \underline{underline}: best heuristic.}
  \label{tab:span_detection}
  \vspace{-1mm}
  \scriptsize
  \begin{tabular}{l cccccc}
    \toprule
    \textbf{Span Method} & \textbf{Prec}$\uparrow$ & \textbf{Rec}$\uparrow$ & \textbf{F1}$\uparrow$ & \textbf{AUROC}$\uparrow$ & \textbf{MAE}$\downarrow$ & $\rho_s\!\uparrow$ \\
    \midrule
    Slide Win (10) & 0.415 & \underline{0.932} & 0.574 & 0.840 & 0.153 & 0.522 \\
    Slide Win (20) & 0.614 & 0.699 & \underline{0.653} & 0.853 & 0.152 & 0.540 \\
    Slide Win (40) & 0.708 & 0.415 & 0.524 & \underline{0.860} & 0.158 & 0.549 \\
    Sent. Boundary  & \underline{0.849} & 0.479 & 0.613 & 0.861 & \underline{0.152} & \underline{0.556} \\
    Token Thr (0.7) & 0.325 & 0.480 & 0.388 & 0.815 & 0.150 & 0.463 \\
    NER (spaCy)     & 0.397 & 0.396 & 0.396 & 0.777 & 0.205 & 0.470 \\
    BIO Tagger      & 0.046 & 0.260 & 0.079 & 0.763 & 0.165 & 0.398 \\
    \midrule
    \ours{} (Ours)  & \textbf{0.857} & \textbf{0.970} & \textbf{0.910} & \textbf{0.927} & \textbf{0.142} & \textbf{0.736} \\
    \bottomrule
  \end{tabular}
  \vspace{-3mm}
\end{wraptable}

\textbf{Detection Quality.}
\ours{}'s DETR detector achieves a detection F1 of \textbf{0.910}, surpassing the best heuristic (Sliding Window 20, F1 = 0.653) by 39.4\%.
It maintains both high precision (0.857) and high recall (0.970), whereas heuristic methods face a precision--recall trade-off: sentence boundaries achieve high precision (0.849) but low recall (0.479), while small sliding windows achieve high recall (0.932) but low precision (0.415).

\textbf{Impact on Uncertainty Estimation.}
The downstream AUROC gap is even more striking: \ours{} achieves \textbf{0.927} versus the best heuristic's 0.861 (+7.7\%).
The span-level correlation $\rho_{\text{span}}$ shows the largest gap: \textbf{0.736} versus 0.556 (+32.4\%), demonstrating that learned span detection is essential for preserving fine-grained uncertainty ordering.

\textbf{BIO Tagger Failure.}
The BIO sequence tagger, the standard approach in span extraction tasks, performs worst (F1 = 0.079, AUROC = 0.763).
This is because BIO tagging produces non-overlapping spans, but ground-truth uncertainty spans frequently overlap (a single token may belong to multiple claims).
The BIO tagger detects 165K candidate spans but only 7{,}663 match ground truth, yielding a precision of merely 4.6\%.
This motivates our DETR-based set prediction approach, which naturally handles overlapping spans via parallel query decoding.

\vspace*{-3mm}
\subsection{Ablation Study}
\label{sec:exp:ablation}
\vspace*{-3mm}

\begin{wraptable}{r}{86mm}
\vspace{-9mm}
  \centering
  \caption{Component ablation. Each row adds one component to the previous configuration. 
  }
  \label{tab:ablation}
  \vspace{-2mm}
  \scriptsize
  \begin{tabular}{l ccccc}
    \toprule
    \textbf{Configuration} & \textbf{AUROC}$\uparrow$ & \textbf{ECE}$\downarrow$ & \textbf{MAE}$\downarrow$ & $\rho_s\!\uparrow$ & $\rho_{\text{seq}}\!\uparrow$ \\
    \midrule
    Base DETR + Single Beta & 0.907 & 0.058 & 0.154 & 0.673 & 0.765 \\
    \quad + Enrichment Gate & 0.920 & 0.052 & 0.144 & 0.725 & 0.729 \\
    \quad + MoB ($K{=}3$)   & 0.933 & 0.020 & 0.116 & 0.760 & 0.846 \\
    \quad + UCIR (1 round)  & \textbf{0.939} & \textbf{0.013} & \textbf{0.106} & \textbf{0.790} & \textbf{0.839} \\
    \bottomrule
  \end{tabular}
  \vspace{-6mm}
\end{wraptable}

We ablate each component of \ours{} by progressively adding modules to the base DETR architecture.
\textbf{Tab.\,\ref{tab:ablation}} reports test-set results; all runs share identical hyperparameters except the ablated component.

\textbf{Enrichment Gate.}
Adding the span enrichment gate improves AUROC by 1.3\% (0.907$\to$0.920) and $\rho_{\text{span}}$ by 5.2\% (0.673$\to$0.725), confirming that providing the uncertainty scorer with direct access to span content is beneficial.
Notably, $\rho_{\text{seq}}$ decreases slightly (0.765$\to$0.729): the enrichment gate sharpens span-level estimates, which can redistribute probability mass among spans within a sequence and introduce minor inconsistencies in the aggregated sequence score.

\textbf{Mixture of Beta (MoB).}
Replacing the single Beta distribution with a $K{=}3$ mixture yields the largest single improvement: +1.3\% AUROC (0.920$\to$0.933), $-$62\% ECE (0.052$\to$0.020), and $-$19\% MAE (0.144$\to$0.116).
The MoB distribution captures the multi-modal nature of uncertainty: certain spans cluster near $u{=}0$, hallucinated spans near $u{=}1$, and partially uncertain spans form an intermediate mode. A single Beta cannot represent this structure.
MoB is also nearly self-calibrated (raw ECE = 0.020, $T{=}1.05$), whereas the single Beta requires aggressive temperature correction ($T{=}0.83$) to reduce its raw ECE from 0.052 to 0.029.
We provide detailed analyses of layer selection strategy and MoB hyperparameters in \textbf{App.\,\ref{app:layer_analysis}} and \textbf{\ref{app:mob_analysis}}, along with a span enrichment analysis (gate vs.\ concat vs.\ add fusion, attention vs.\ mean pooling) in \textbf{App.\,\ref{app:enrichment_analysis}}.

\textbf{Uncertainty-Conditioned Iterative Refinement (UCIR).}
A single UCIR round provides the second-largest improvement: +0.6\% AUROC (0.933$\to$0.939) and +3.0\% $\rho_{\text{span}}$ (0.760$\to$0.790).
UCIR allows the model to refine its uncertainty estimates by conditioning on its own initial predictions, effectively implementing a ``second look'' that corrects systematic errors.
The shared decoder reuses detection weights with only a lightweight MLP adapter and residual connection ($\alpha{=}0.7$), amounting to negligible parameter overhead.
The improvement is most pronounced for $\rho_{\text{span}}$, suggesting that UCIR primarily improves the \emph{ranking} of uncertainty estimates: conditioning on the initial estimate allows the model to identify and correct systematic biases, e.g., spans initially predicted as moderately uncertain but that should be highly uncertain based on their content.
The marginal $\rho_{\text{seq}}$ decrease (0.846$\to$0.839) mirrors the enrichment gate pattern: span-focused refinement slightly perturbs the per-span scores that feed into sequence-level aggregation, but the effect is small ($<$1\%) and is offset by substantial gains on all other metrics.

\begin{wraptable}{r}{85mm}
\vspace{-4mm}
  \centering
  \caption{Span boundary prediction: regression vs. \ pointer. Same DETR architecture (nq=32, enc2+dec3).
  }
  \label{tab:regression}
  \vspace{-2mm}
  \scriptsize
  \begin{tabular}{l ccccc}
    \toprule
    \textbf{Method} & \textbf{AUROC}$\uparrow$ & \textbf{MAE}$\downarrow$ & \textbf{Prec}$\uparrow$ & \textbf{Rec}$\uparrow$ & $\rho_s\!\uparrow$ \\
    \midrule
    Pointer (cross-attn) & 0.888 & 0.167 & 0.767 & 0.970 & ---\textsuperscript{$\dagger$} \\
    Regression (L1+GIoU) & \textbf{0.922} & \textbf{0.143} & \textbf{0.838} & \textbf{0.970} & \textbf{0.725} \\
    \bottomrule
  \end{tabular}
  \vspace{-3mm}
\end{wraptable}

\textbf{Regression vs.\ Pointer for Span Detection.} \textbf{Tab.\,\ref{tab:regression}} compares our continuous regression head against the more common cross-attention pointer mechanism.
Regression outperforms the pointer by 3.4\% AUROC and 7.1\% precision while maintaining identical recall (0.970).
The pointer mechanism suffers from discretization artifacts: argmax over cross-attention weights cannot represent fractional positions and struggles with long spans where attention is diffuse.
Regression directly predicts continuous start/end coordinates, enabling smoother gradient flow and more precise boundary localization.
Additional analyses on robustness across hyperparameter configurations and sensitivity to individual hyperparameters are in \textbf{App.\,\ref{app:robustness}} and \textbf{\ref{app:hp_sensitivity}}.

\section{Conclusion}
\label{sec:conclusion}
\vspace*{-4mm}

We formalized Span-Level Uncertainty Estimation (SLUE) and introduced \ours{}, a lightweight (${\sim}$25M-parameter) probe that distills multi-sample uncertainty into a single forward pass over frozen LLM hidden states, combining a DETR-style span decoder, Mixture of Beta head, and Uncertainty-Conditioned Iterative Refinement.
Together with \bench{}, the first span-level uncertainty benchmark (293K spans, five domains), experiments on five LLMs (4B--30B, dense and MoE) show that \ours{} achieves AUROC 0.908--0.944 and MAE 0.110--0.129 while being $10$--$20\times$ faster than sampling-based methods.
The DETR detector attains 0.910 F1, surpassing the best heuristic by 39.4\%, and the learned span-to-sequence composition ($\rho_{\text{seq}} = 0.839$) suggests that span-level estimation subsumes sequence-level as a special case.

\textbf{Future directions.}
Because span-level scores are attached to individual claims, they can serve as a decision signal for downstream systems rather than only an evaluation metric.
We see three directions this enables:
(i) \emph{targeted verification}, which spends retrieval or stronger-model checks only on high-uncertainty spans;
(ii) \emph{span-level self-refinement}, which revises only the spans that need it instead of regenerating the whole response; and
(iii) \emph{uncertainty-aware multi-agent communication}, where agents pass claim-level reliability signals along with their text.
We discuss these directions and their open problems in \textbf{App.~\ref{app:future_directions}}.

\textbf{Limitations and Broader Impacts.}
\ours{} cannot capture uncertainty from the generation process itself
and requires white-box access to hidden states.
The multi-sample label construction pipeline is compute-intensive, though this is a one-time cost.
Our evaluation focuses on claim uncertainty in English; extending to other languages and uncertainty types (e.g., reasoning) remains future work.
%
Our labels quantify \emph{cross-sample consistency}: $u^*$ is the contradiction rate among a model's own samples that take a position on a claim, with no retrieval or external knowledge source entering the computation (\textbf{App.\,\ref{app:distillation}}).
It is therefore not a factuality label, and the two can diverge: 22.3\% of high-$u^*$ spans are factually correct, though an audit attributes ${\sim}$89\% of those to genuine cross-sample disagreement rather than pipeline defects (\textbf{App.\,\ref{app:fp_audit}}).
Such labels are also blind by construction to \emph{confident} hallucinations that recur identically across samples.
Our evidence for label validity  
(human agreement, monotone risk reduction, and downstream selective prediction)
is end-to-end; we do not ablate the pipeline component-wise, so the residual noise contributed individually by claim decomposition, verification, and span alignment remains unquantified, and characterizing it is the natural next step for \bench{}. We discuss societal benefits and risks, including the danger of overreliance on unflagged spans, in \textbf{App.\,\ref{app:broader_impacts}}.


\section*{Acknowledgements and Disclosures}
This work was supported by Amazon through computational resources and access to large language model services.
The authors declare no competing interests.

\bibliographystyle{IEEEtranN}
\bibliography{refs/ref}

\clearpage\newpage
\onecolumn
\section*{\Large{Appendix}}
\setcounter{section}{0}
\setcounter{figure}{0}
\setcounter{table}{0}
\makeatletter 
\renewcommand{\thesection}{\Alph{section}}
\renewcommand{\theHsection}{\Alph{section}}
\renewcommand{\thefigure}{A\arabic{figure}}
\renewcommand{\theHfigure}{A\arabic{figure}}
\renewcommand{\thetable}{A\arabic{table}}
\renewcommand{\theHtable}{A\arabic{table}}
\makeatother

\section{Architectural Design Details of \ours{}}
\label{app:span_query_lifecycle}

\begin{figure*}[htb]
\centering
\resizebox{\textwidth}{!}{%
\begin{tikzpicture}[
    >=Stealth,
    box/.style={draw, rounded corners=2pt, minimum height=0.65cm, align=center, font=\small},
    querybox/.style={draw, rounded corners=2pt, fill=green!8, draw=green!50!black,
                     minimum height=0.5cm, align=center, font=\small},
    tokenbox/.style={draw, rounded corners=1pt, fill=blue!6, draw=blue!30,
                     minimum height=0.45cm, minimum width=0.65cm, align=center, font=\scriptsize},
    headbox/.style={draw, rounded corners=2pt, minimum height=0.5cm, align=center, font=\small},
    stagelabel/.style={font=\small\bfseries\sffamily, text=black!70},
    annot/.style={font=\scriptsize, text=black!60},
    darrow/.style={->, thick, draw=black!60},
    every node/.append style={inner sep=3.5pt},
]



\node[stagelabel] at (2.5, 11.0) {Stage 1: Initialization};

\node[box, fill=yellow!10, draw=yellow!50!black, minimum width=3.0cm, font=\scriptsize]
    (embed) at (2.5, 10.2) {\texttt{nn.Embedding($N$, $d_p$)}};

\node[querybox, minimum width=1.4cm, font=\scriptsize, fill=red!8, draw=red!30] (q1) at (0.3, 9.0) {$\vq_1$};
\node[querybox, minimum width=1.4cm, font=\scriptsize, fill=orange!10, draw=orange!40] (q2) at (1.8, 9.0) {$\vq_2$};
\node[querybox, minimum width=1.4cm, font=\scriptsize, fill=blue!8, draw=blue!30] (q3) at (3.2, 9.0) {$\vq_3$};
\node[font=\normalsize, text=black!40] at (4.1, 9.0) {$\cdots$};
\node[querybox, minimum width=1.4cm, fill=violet!8, draw=violet!30, font=\scriptsize] (qN) at (5.0, 9.0) {$\vq_{32}$};

\draw[darrow] (embed) -- (2.5, 9.4);

\node[annot, text width=7.0cm, align=center] at (2.5, 8.2)
    {$N{=}32$ content-agnostic ``span prototypes'', each $\vq_k \in \mathbb{R}^{d_p}$\\($d_p{=}512$)};

\node[stagelabel] at (14.0, 11.0) {Stage 2: Token Encoder};

\draw[draw=teal!50, thick, rounded corners=6pt, fill=teal!3]
    (8.0, 8.6) rectangle (20.0, 10.7);
\node[stagelabel, text=teal!60!black, font=\scriptsize\bfseries] at (14.0, 10.45)
    {Multi-Layer Hidden-State Fusion};

\node[box, fill=teal!8, draw=teal!40, minimum width=4.5cm, font=\scriptsize] (hid) at (10.8, 9.7)
    {LLM Hidden States $\mH^{(\ell)}$};
\node[annot, text width=3.5cm, align=center] at (10.8, 9.15)
    {\textit{Layers 22, 24, 26}\\(mean fusion)};

\node[box, fill=teal!12, draw=teal!50, minimum width=3.5cm, font=\scriptsize] (proj) at (15.5, 9.7)
    {Linear Proj $\to d_p$};
\node[annot, text width=3.0cm, align=center] at (15.5, 9.15)
    {\textit{+ token entropy}\\as extra feature};

\draw[darrow] (hid.east) -- (proj.west);

\node[annot, text=teal!60!black, font=\scriptsize\bfseries] at (18.5, 9.7) {per token};

\node[tokenbox, fill=blue!5] (t0) at (10.2, 7.9) {\strut was};
\node[tokenbox, fill=blue!5] (t1) at (10.9, 7.9) {\strut born};
\node[tokenbox, fill=blue!5] (t2) at (11.6, 7.9) {\strut as};
\node[tokenbox, fill=orange!18, draw=orange!50] (t3) at (12.3, 7.9) {\strut Derek};
\node[tokenbox, fill=orange!18, draw=orange!50] (t4) at (13.15, 7.9) {\strut Dele..};
\node[tokenbox, fill=orange!18, draw=orange!50] (t5) at (13.95, 7.9) {\strut Harris};
\node[tokenbox, fill=blue!5] (t6) at (14.65, 7.9) {\strut on};
\node[tokenbox, fill=blue!5] (t7) at (15.25, 7.9) {\strut Aug};
\node[tokenbox, fill=orange!18, draw=orange!50] (t8) at (15.85, 7.9) {\strut 12};
\node[tokenbox, fill=orange!18, draw=orange!50] (t9) at (16.4, 7.9) {\strut ,};
\node[tokenbox, fill=orange!18, draw=orange!50] (t10) at (16.95, 7.9) {\strut 1926};
\node[tokenbox, fill=blue!5] (t11) at (17.6, 7.9) {\strut ...};

\node[annot, font=\scriptsize\bfseries, text=teal!70] at (14.0, 7.35)
    {Output: $\mZ \in \mathbb{R}^{T \times d_p}$ \;(token features)};

\draw[darrow, thick, draw=teal!60] (14.0, 8.6) -- (14.0, 8.2);

\draw[dashed, black!20, line width=0.8pt] (-1.5, 6.8) -- (20.5, 6.8);

\node[stagelabel] at (6.0, 6.45) {Stage 3: DETR-Style Span Decoder};

\draw[draw=green!50!black, thick, rounded corners=6pt, fill=green!3]
    (1.5, 4.6) rectangle (19.0, 6.1);

\node[box, fill=green!12, draw=green!40, minimum width=2.8cm, font=\scriptsize] (sa) at (4.5, 5.2)
    {Self-Attention};
\node[annot, text width=3.5cm, align=center] at (4.5, 4.25)
    {\textit{Inter-span dependencies}\\queries attend to each other};

\node[box, fill=teal!12, draw=teal!50, minimum width=3.8cm, font=\scriptsize] (ca) at (10.25, 5.2)
    {Cross-Attention};
\node[annot, text width=3.5cm, align=center] at (10.25, 4.25)
    {\textit{Q}=$\mQ$, \textit{K,V}=$\mZ$\\localize span positions};

\node[box, fill=gray!8, draw=gray!40, minimum width=3.8cm, font=\scriptsize] (ffn) at (16.0, 5.2)
    {FFN + LayerNorm};

\node[annot, text=green!50!black, font=\scriptsize\bfseries] at (18.5, 5.2) {$\times 3$};

\draw[darrow] (sa.east) -- (ca.west);
\draw[darrow] (ca.east) -- (ffn.west);

\draw[darrow, dashed, draw=gray!50, thin, rounded corners=3pt]
    (sa.north) -- ++(0, 0.25) -| ([xshift=-25pt]ca.north)
    node[pos=0.15, above=1pt, font=\tiny, text=gray!60] {+residual};
\draw[darrow, dashed, draw=gray!50, thin, rounded corners=3pt]
    ([xshift=-25pt]ca.north) -- ++(0, 0.25) -| (ffn.north);


\draw[darrow, thick, line width=1.2pt, draw=green!60!black]
    (2.5, 7.8) -- (2.5, 5.5) |- (sa.west)
    node[pos=0.15, right=2pt, annot, text=green!50!black] {$\mQ$};

\draw[thick, line width=1.2pt, draw=teal!60]
    (14.0, 7.2) -- (14.0, 6.5)
    node[pos=0.0, right=2pt, annot, text=teal!60]{} ;
\draw[thick, line width=1.2pt, draw=teal!60]
    (10.25, 6.5) -- (19.5, 6.5);
\draw[darrow, thick, line width=1.2pt, draw=teal!60]
    (10.25, 6.5) -- (ca.north);
\draw[darrow, thick, line width=1.2pt, draw=teal!60]
    (19.5, 6.5) -- (19.5, 2.0) -| ([xshift=2.6cm, yshift=1.15cm] enrich.north east);

\node[annot, text=orange!70, font=\scriptsize\itshape] at (10.25, 3.7)
    {e.g.\ $\hat{\vq}_2$ attends to ``Derek Delevan Harris'' tokens via cross-attn};

\node[querybox, minimum width=8.0cm, fill=green!18, draw=green!60!black] (qhat) at (10.25, 3.2)
    {$\hat{\mQ} \in \mathbb{R}^{32 \times d_p}$ \;\;(position-aware refined queries)};

\draw[darrow, thick, draw=green!60!black, line width=1.0pt]
  (16.0, 4.6) -- (16.0, 3.2) -- (qhat.east);

\draw[dashed, black!20, line width=0.8pt] (-1.5, 2.6) -- (20.5, 2.6);

\node[stagelabel] at (5.5, 2.3) {Stage 4: Prediction Heads};


\coordinate (qfork) at (10.25, 2.0);
\draw[thick, draw=green!60!black, line width=1.2pt] (qhat.south) -- (qfork)
    node[midway, right=2pt, annot, text=green!50!black] {$\hat{\mQ}$};
\draw[darrow, thick, draw=green!60!black, line width=1.0pt] (qfork) -| (4.5, 1.52);
\draw[darrow, thick, draw=green!60!black, line width=1.0pt] (qfork) -| (15.0, 1.52);

\node[headbox, fill=orange!10, draw=orange!50, minimum width=8.5cm, minimum height=0.6cm]
    (bhead) at (4.5, 1.2) {\textbf{Boundary Regression}};

\node[annot, text width=8.0cm, align=left] at (4.5, 0.2)
    {$\hat{\vq}_k \xrightarrow{\;\text{MLP}\;} \sigma(\cdot) \;\to\; (\hat{b}_k,\, \hat{e}_k) \in [0,1]^2$\\[3pt]
     \textcolor{orange!70!black}{Ex:} $\hat{\vq}_2 \to (\hat{b}_2{=}0.18,\, \hat{e}_2{=}0.31)$
     \;\;\footnotesize Maps to ``\textit{Derek Delevan Harris}''};

\node[headbox, fill=yellow!12, draw=yellow!60!black, minimum width=9.5cm, minimum height=0.6cm]
    (enrich) at (15.0, 1.2) {\textbf{Span Enrichment Gate}};

\node[annot, text width=4.5cm, align=left, anchor=north east] at (15.55, 0.6)
    {Soft boundary mask:\\[2pt]
     $m_k(t){=}\sigma(\tau(t{-}\hat{b}_k))\!\cdot\!\sigma(\tau(\hat{e}_k{-}t))$\\[4pt]
     \begin{tikzpicture}[baseline=-0.3em, scale=0.6]
       \fill[blue!8] (0,0) rectangle (0.35,0.08);
       \fill[blue!8] (0.4,0) rectangle (0.75,0.12);
       \fill[orange!30] (0.8,0) rectangle (1.15,0.7);
       \fill[orange!50] (1.2,0) rectangle (1.55,0.95);
       \fill[orange!40] (1.6,0) rectangle (1.95,0.8);
       \fill[blue!8] (2.0,0) rectangle (2.35,0.1);
       \fill[blue!8] (2.4,0) rectangle (2.75,0.06);
       \node[font=\tiny, text=orange!70!black, rotate=90] at (-0.25, 0.5) {$m_k$};
     \end{tikzpicture}};

\node[annot, text width=4.5cm, align=left, anchor=north west] at (15.15, 0.6)
    {Mask-weighted attn-pool $\to \vc_k$\\[3pt]
     Enriched queries: $\tilde{\vq}_k = g_k \vc_k + (1{-}g_k)\hat{\vq}_k$\\[3pt]
     \textcolor{yellow!60!black}{Ex:} $g_2{=}0.72$ (content-heavy)};

\draw[darrow, thick, draw=orange!60] (bhead.east) -- (enrich.west)
    node[midway, above=2pt, annot, text=orange!70] {$(\hat{b}_k, \hat{e}_k)$};



\node[headbox, fill=cyan!10, draw=cyan!50, minimum width=8.5cm, minimum height=0.6cm]
    (vhead) at (4.5, -1.8) {\textbf{Validity Classification}};

\node[annot, text width=8.0cm, align=left] at (4.5, -2.8)
    {$\tilde{\vq}_k \xrightarrow{\;\text{MLP}\;} \sigma(\cdot) \;\to\; p_k^{\text{valid}}$\\[3pt]
     \textcolor{green!60!black}{\ding{51}} $\tilde{\vq}_2 \to 0.97$ (real span)\quad
     \textcolor{red!60}{\ding{55}} $\tilde{\vq}_{28} \to 0.03$ (no span)};

\node[headbox, fill=purple!10, draw=purple!50, minimum width=9.5cm, minimum height=0.6cm]
    (uhead) at (15.0, -1.8) {\textbf{MoB Uncertainty}};

\node[annot, text width=4.2cm, align=left, anchor=north west] at (10.5, -2.25)
    {$\tilde{\vq}_k \xrightarrow{\;\text{MLP}\;} \{(\pi_j, \alpha_j, \beta_j)\}_{j=1}^{K}$\\[6pt]
     $\hat{u}_k = \textstyle\sum_j \pi_j \frac{\alpha_j}{\alpha_j + \beta_j}$\\[4pt]
     \textcolor{purple!70}{e.g.\ $\hat{u}_2{=}0.94$}};

\begin{scope}[shift={(17.0, -3.7)}, scale=1.1]
  \draw[thick, red!60, domain=0:1, samples=40] plot (\x, {1.2*\x*\x*\x*\x});
  \draw[thick, blue!60, domain=0:1, samples=40] plot (\x, {2.5*\x*(1-\x)});
  \draw[thick, green!50!black, domain=0:1, samples=40] plot (\x, {1.8*\x*\x*\x*(1-\x)});
  \draw[very thick, purple!70, dashed, domain=0:1, samples=40]
      plot (\x, {0.5*1.2*\x*\x*\x*\x + 0.3*2.5*\x*(1-\x) + 0.2*1.8*\x*\x*\x*(1-\x)});
  \draw[->] (0,0) -- (1.2,0) node[right, font=\scriptsize] {$u$};
  \draw[->] (0,0) -- (0,1.15);
  \node[draw=gray!40, fill=white, rounded corners=2pt, inner sep=3pt,
        font=\tiny, anchor=north west, align=left] at (-1.2, 1)
    {\textcolor{red!60}{\rule{6pt}{1.5pt}} $\;\mathrm{Beta}_1$\\
     \textcolor{blue!60}{\rule{6pt}{1.5pt}} $\;\mathrm{Beta}_2$\\
     \textcolor{green!50!black}{\rule{6pt}{1.5pt}} $\;\mathrm{Beta}_3$\\
     \textcolor{purple!70}{\rule[0.5ex]{3pt}{0.8pt}\rule[0.5ex]{1pt}{0pt}\rule[0.5ex]{3pt}{0.8pt}} $\;\mathrm{Mix}$};
\end{scope};

\draw[darrow, thick, draw=yellow!60!black] (enrich.south west) -- (vhead.north east)
    node[pos=0.4, left=2pt, annot, text=yellow!60!black] {$\tilde{\vq}_k$};

\draw[darrow, thick, draw=yellow!60!black] (enrich.south) -- ++(0, -1.0) -- (uhead.north)
    node[pos=0.3, right=2pt, annot, text=yellow!60!black, xshift=3pt, yshift=-15pt] {$\tilde{\vq}_k$};


\draw[dashed, black!20, line width=0.8pt] (-1.5, -4.1) -- (20.5, -4.1);

\node[stagelabel] at (10.0, -4.5) {Stage 5: Output};

\node[annot, font=\small, text=black, text width=19cm, align=left] at (10.0, -5.4)
    {\textit{``John Derek was}
     \colorbox{red!15}{\textit{born as Derek Delevan Harris}}
     \textit{on} \colorbox{red!15}{\textit{August 12, 1926}}\textit{,}
     \textit{in} \colorbox{green!15}{\textit{Hollywood, California}}\textit{.}
     \textit{He was the}
     \colorbox{red!15}{\textit{son of a director}}
     \textit{and}
     \colorbox{red!15}{\textit{won the Golden Globe}}
     \textit{for his debut film ...''}
     \quad
     \colorbox{red!15}{\scriptsize high $\hat{u}$: hallucinated}\;
     \colorbox{green!15}{\scriptsize low $\hat{u}$: factual}};

\draw[rounded corners=3pt, fill=gray!6, draw=gray!20, dashed]
    (0.5, -9.3) rectangle (7.5, -6.5);
\node[annot, text width=6.0cm, align=center, font=\scriptsize] at (4.0, -6.9)
    {\textbf{Filtered out} (validity $< 0.5$):};
\node[annot, text width=6.0cm, align=center] at (4.0, -7.5)
    {$\hat{\vq}_1, \hat{\vq}_3, \hat{\vq}_4, \hat{\vq}_5, \hat{\vq}_{20}, \ldots, \hat{\vq}_{32}$\\[2pt]
     27 of 32 queries $\to$ \textcolor{gray}{no span detected}};

\draw[rounded corners=3pt, fill=purple!5, draw=purple!30]
    (0.5, -8.0) rectangle (7.5, -9.3);
\node[annot, text width=6.0cm, align=center, font=\scriptsize] at (4.0, -8.25)
    {\textbf{Sequence-Level Aggregation}};
\node[annot, text width=6.0cm, align=center] at (4.0, -8.85)
    {$\hat{u}_{\text{seq}} = \frac{\sum_k p_k^{\text{valid}} \cdot \hat{u}_k}{\sum_k p_k^{\text{valid}}}$
     \;\;\textcolor{purple!70}{$= 0.81$}};

\node[font=\scriptsize\bfseries] at (14.0, -6.2) {Detected Spans (validity $> 0.5$):};

\draw[rounded corners=3pt, fill=gray!3, draw=gray!25]
    (8.0, -9.3) rectangle (20.0, -6.5);

\node[font=\scriptsize\bfseries, anchor=west] at (8.2, -6.75) {Query};
\node[font=\scriptsize\bfseries, anchor=west] at (9.3, -6.75) {Detected Span};
\node[font=\scriptsize\bfseries] at (14.0, -6.75) {Valid};
\node[font=\scriptsize\bfseries] at (15.1, -6.75) {$\hat{u}$};
\node[font=\scriptsize\bfseries] at (16.0, -6.75) {$u^*$};
\node[font=\scriptsize\bfseries, anchor=west] at (16.8, -6.75) {Verdict};
\draw[gray!30] (8.0, -6.95) -- (20.0, -6.95);

\node[font=\scriptsize, anchor=west] at (8.2, -7.25) {$\hat{\vq}_2$};
\node[font=\scriptsize, anchor=west, text=red!70!black] at (9.3, -7.25) {Derek Delevan Harris};
\node[font=\scriptsize] at (14.0, -7.25) {0.97};
\node[font=\scriptsize, text=red!70!black] at (15.1, -7.25) {0.94};
\node[font=\scriptsize] at (16.0, -7.25) {1.0};
\node[font=\scriptsize, text=red!70!black, anchor=west] at (16.8, -7.25) {\ding{55} hallucinated};

\node[font=\scriptsize, anchor=west] at (8.2, -7.65) {$\hat{\vq}_7$};
\node[font=\scriptsize, anchor=west, text=red!70!black] at (9.3, -7.65) {August 12, 1926};
\node[font=\scriptsize] at (14.0, -7.65) {0.95};
\node[font=\scriptsize, text=red!70!black] at (15.1, -7.65) {0.89};
\node[font=\scriptsize] at (16.0, -7.65) {0.95};
\node[font=\scriptsize, text=red!70!black, anchor=west] at (16.8, -7.65) {\ding{55} hallucinated};

\node[font=\scriptsize, anchor=west] at (8.2, -8.05) {$\hat{\vq}_{11}$};
\node[font=\scriptsize, anchor=west, text=green!50!black] at (9.3, -8.05) {Hollywood, California};
\node[font=\scriptsize] at (14.0, -8.05) {0.93};
\node[font=\scriptsize, text=green!50!black] at (15.1, -8.05) {0.03};
\node[font=\scriptsize] at (16.0, -8.05) {0.0};
\node[font=\scriptsize, text=green!50!black, anchor=west] at (16.8, -8.05) {\ding{51} factual};

\node[font=\scriptsize, anchor=west] at (8.2, -8.45) {$\hat{\vq}_{15}$};
\node[font=\scriptsize, anchor=west, text=red!70!black] at (9.3, -8.45) {won the Golden Globe};
\node[font=\scriptsize] at (14.0, -8.45) {0.91};
\node[font=\scriptsize, text=red!70!black] at (15.1, -8.45) {0.95};
\node[font=\scriptsize] at (16.0, -8.45) {1.0};
\node[font=\scriptsize, text=red!70!black, anchor=west] at (16.8, -8.45) {\ding{55} hallucinated};

\node[font=\scriptsize, anchor=west] at (8.2, -8.85) {$\hat{\vq}_{19}$};
\node[font=\scriptsize, anchor=west, text=red!70!black] at (9.3, -8.85) {son of a director};
\node[font=\scriptsize] at (14.0, -8.85) {0.88};
\node[font=\scriptsize, text=red!70!black] at (15.1, -8.85) {0.95};
\node[font=\scriptsize] at (16.0, -8.85) {1.0};
\node[font=\scriptsize, text=red!70!black, anchor=west] at (16.8, -8.85) {\ding{55} hallucinated};

\draw[darrow, dashed, thick, draw=purple!50, line width=1.2pt, rounded corners=8pt]
    (20.0, -7.8) -- (20.3, -7.8) -- (20.3, 5.5) -- (19.0, 5.5)
    node[pos=0.35, right=3pt, annot, text=purple!60, font=\scriptsize\bfseries, text width=1.5cm, align=center]
    {UCIR\\Round 2};

\end{tikzpicture}%
}
\caption{
\textbf{Span query lifecycle in \ours{}.}
\textbf{Stage~1}: $N{=}32$ learnable span queries are initialized as content-agnostic prototypes.
\textbf{Stage~2}: A token encoder fuses multi-layer hidden states with token entropy, producing the token feature pool $\mZ \in \mathbb{R}^{T \times d_p}$.
\textbf{Stage~3}: A 3-layer DETR decoder refines queries via self-attention (inter-span dependencies) and cross-attention over $\mZ$ (span localization), yielding position-aware $\hat{\mQ}$.
$\mZ$ is shared between cross-attention and span enrichment (T-shaped data flow on the right).
\textbf{Stage~4}: Four prediction heads operate on each query: \emph{boundary regression} predicts start/end positions $(\hat{b}_k, \hat{e}_k)$; \emph{span enrichment} applies a differentiable soft mask to attention-pool $\mZ$ into a content vector, fused with the query via a learned gate to produce $\tilde{\vq}_k$; \emph{validity classification} scores whether the query corresponds to a real span; \emph{MoB uncertainty} outputs $K{=}3$ Beta components whose weighted mean gives $\hat{u}_k$.
\textbf{Stage~5}: Valid spans (score ${>}\,0.5$) are retained and aggregated into a sequence-level uncertainty via importance-weighted averaging.
The dashed purple arrow indicates the UCIR feedback loop (Sec.~\ref{sec:method:inference}).
The example shows a biography of John Derek: \ours{} assigns high uncertainty ($\hat{u} \geq 0.89$) to hallucinated claims and low uncertainty ($\hat{u} = 0.03$) to the factual birthplace.
}
\label{fig:span_query_lifecycle}
\end{figure*}
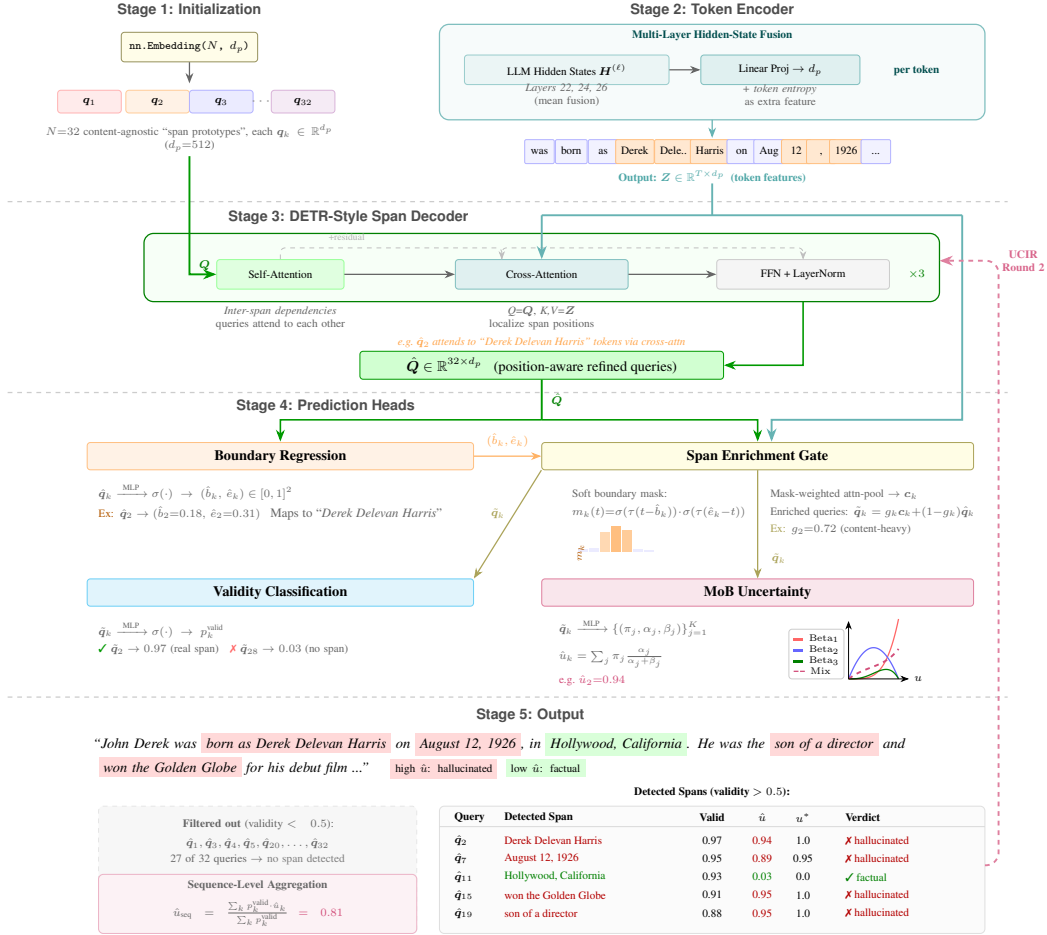

\textbf{Fig.\,\ref{fig:span_query_lifecycle}} illustrates the end-to-end span-level uncertainty estimation process in \ours{}, from initialization to final uncertainty output.
This appendix expands on the architectural overview in Sec.\,\ref{sec:method:arch} by providing design rationale and implementation details for each stage.

\subsection{Stage 1: Span Query Initialization}
\label{app:lifecycle:init}

\ours{} initializes $N$ learnable span queries $\mQ \in \sR^{N \times d_p}$ as content-agnostic embedding vectors (Xavier uniform).
At the start of training, no query ``knows'' which span it will eventually detect; specialization emerges entirely through learning.

\paragraph{Why learnable queries instead of input-derived proposals?}
An alternative design would derive candidate spans from the input, e.g., via a sliding window, constituency parse, or BIO sequence tagger.
We deliberately avoid this for three reasons.
First, input-derived proposals impose a fixed segmentation before the model has seen the full context, which precludes overlapping spans and makes the detector sensitive to segmentation errors.
Second, a BIO tagger produces spans sequentially and cannot model inter-span dependencies in parallel.
Third, learnable queries decouple the \emph{number} of detection slots from the input length, keeping the decoder cost constant regardless of sequence length.

\paragraph{Choosing $N$.}
The number of queries $N$ sets an upper bound on how many spans can be detected per response.
In our benchmark, the average response contains ${\sim}$15 spans (31{,}332 test spans across 2{,}000 queries for Qwen3-14B), with a 95th-percentile of ${\sim}$28.
We set $N{=}32$ to comfortably cover the tail while keeping the decoder's $O(N^2)$ self-attention cost negligible.
A hyperparameter sweep (Tab.\,\ref{tab:hp_sensitivity}) confirms that $N{=}16$ already achieves comparable AUROC (0.905 vs.\ 0.910 for $N{=}32$), indicating that the model gracefully degrades when a few spans exceed the slot budget.
Queries that do not bind to any span are suppressed by the validity head (Sec.\,\ref{app:lifecycle:heads}).

\paragraph{Content-agnostic initialization.}
Because all queries start from the same random distribution, no query is biased toward a particular position or span type.
During training, the combination of Hungarian matching (which assigns each ground-truth span to the best-matching query) and decoder self-attention (which encourages queries to diversify) drives the queries to specialize.
Empirically, we observe that after training, different queries tend to cover different positional ranges of the response, though this specialization is soft and adaptive rather than hard-coded.

\subsection{Stage 2: Token Encoder and Feature Pool}
\label{app:lifecycle:encoder}

Before the span decoder can attend to the response, the raw LLM hidden states must be compressed and contextualized into a shared feature pool $\mZ$ (Eq.\,\ref{eq:encoder}).

\paragraph{Multi-layer fusion.}
Different layers of a Transformer LLM encode qualitatively different information: early layers capture syntax and positional structure, middle layers encode factual associations, and late layers specialize in next-token prediction~\citep{orgad2024llms}.
Uncertainty signals are distributed across this hierarchy.
For instance, a factually incorrect span may appear confident in the final layer (high next-token probability) yet show conflicting activations in mid-layers where factual recall occurs.
By fusing layers from the peak uncertainty zone (identified via single-layer probing; see App.\,\ref{app:layer_analysis}), we capture complementary signals that no single layer provides.
We use simple mean fusion (Eq.\,\ref{eq:fusion}) rather than learned weighted fusion or concatenation-projection.
Our ablation (Tab.\,\ref{tab:layer}) shows that mean fusion matches or exceeds learned alternatives while adding zero parameters, likely because the three selected layers are already close in representation space and a learned weighting offers little additional expressiveness.

\paragraph{Projection and contextualization.}
The fused hidden states $\bar{\mH} \in \sR^{T \times d}$ are linearly projected from the LLM dimension $d$ (e.g., 5{,}120 for Qwen3-14B) to a much smaller $d_p{=}512$.
This 10$\times$ compression is critical: it reduces the decoder's memory and compute cost quadratically, and acts as an information bottleneck that forces the model to retain only uncertainty-relevant features.
The projected tokens are then passed through a 2-layer Transformer encoder with sinusoidal positional encodings.
This encoder serves two purposes: (i)~it re-contextualizes the projected tokens so that each position is aware of its neighbors (important because the LLM's own positional information is partially lost after projection), and (ii)~it provides a non-linear transformation that adapts the frozen LLM features to the uncertainty estimation task.

\paragraph{The shared feature pool $\mZ$.}
The encoder output $\mZ \in \sR^{T \times d_p}$ serves as the shared ``memory'' for two downstream consumers: the decoder's cross-attention (Stage~3) and the span feature enrichment module (Stage~4).
This T-shaped data flow is deliberate.
Sharing $\mZ$ ensures that the boundary predictions (which depend on cross-attention over $\mZ$) and the content features (which are pooled from $\mZ$) operate in the same representation space, avoiding the need for an additional alignment step.
It also means that gradients from both the boundary loss and the uncertainty loss flow back through the same encoder, encouraging $\mZ$ to encode both positional and semantic information.

\subsection{Stage 3: DETR-Style Span Decoder}
\label{app:lifecycle:decoder}

The decoder is the core of \ours{}'s detection mechanism.
It transforms the $N$ content-agnostic queries into $N$ content-aware span representations $\hat{\mQ}$ via iterative cross-attention over $\mZ$ (Eq.\,\ref{eq:decoder}).

\paragraph{Decoder architecture.}
Each of the 3 decoder layers applies, in order: (i)~multi-head self-attention among the $N$ queries, (ii)~multi-head cross-attention from queries to the token pool $\mZ$, and (iii)~a position-wise feed-forward network.
Layer normalization is applied before each sub-layer (pre-norm), and residual connections wrap each sub-layer.

\paragraph{Self-attention: inter-query coordination.}
Self-attention allows queries to ``see'' each other.
This serves two functions.
First, it enables \emph{competition}: if two queries attend to the same region of $\mZ$, self-attention lets them negotiate so that one shifts to a different span, reducing duplicate detections.
Second, it enables \emph{cooperation}: queries covering adjacent or overlapping spans can share contextual information, which is useful when a long sentence contains multiple interleaved claims.
Without self-attention, each query would operate in isolation, and the model would need to rely entirely on the Hungarian matching loss to discourage duplicates, a much weaker signal.

\paragraph{Cross-attention: binding queries to spans.}
Cross-attention is the mechanism by which each query ``finds'' its span in the token sequence.
Each query computes attention weights over all $T$ token positions in $\mZ$, and the resulting weighted sum becomes the query's updated representation.
Over multiple decoder layers, the attention pattern sharpens: in early layers, a query may attend broadly to a sentence-level region; by the final layer, attention concentrates on the specific tokens that constitute the span.
This progressive refinement is analogous to how DETR's object queries gradually localize from coarse regions to tight bounding boxes.

\paragraph{Why 3 decoder layers?}
We follow the DETR convention of using a small number of decoder layers (the original DETR uses 6 for the more complex 2D object detection task).
Our ablation shows that 2 layers underperform (AUROC 0.901 vs.\ 0.910 for 3 layers), while 4 layers provide no further gain, suggesting that 3 layers suffice for the 1D span detection problem.
Each additional layer adds cross-attention and self-attention, so the cost grows linearly in the number of layers.

\paragraph{Handling overlapping spans.}
Unlike BIO-based sequence taggers, which assign exactly one label per token and therefore cannot represent overlapping spans, the DETR formulation places no constraint on the spatial relationship between queries.
Two queries can predict spans that partially or fully overlap, because each query independently regresses its own $(b_k, e_k)$ boundaries.
This is important in practice: for example, in the sentence ``Marie Curie was a Polish-French physicist who won the Nobel Prize in Physics in 1903,'' the spans ``Polish-French physicist'' and ``the Nobel Prize in Physics'' may share boundary tokens depending on the granularity of claim decomposition.

\paragraph{Auxiliary losses on intermediate layers.}
Following DETR, we apply the same prediction heads and losses to the output of each intermediate decoder layer (not just the final layer), with a reduced weight $\lambda_{\text{aux}}{=}0.5$.
This provides direct supervision to early layers, accelerating convergence and improving gradient flow through the deep decoder stack.
Without auxiliary losses, the early decoder layers receive only indirect gradients through the final layer, which can slow training significantly.

\subsection{Stage 4: Prediction Heads and Span Enrichment}
\label{app:lifecycle:heads}

After decoding, each query $\hat{\vq}_k$ is passed through four prediction heads.
The design of these heads reflects a careful separation of concerns: boundary prediction uses the \emph{pre-enrichment} query, while uncertainty and validity prediction use the \emph{enriched} query.

\paragraph{Boundary regression.}
The boundary head is a 3-layer MLP that maps $\hat{\vq}_k$ to normalized coordinates $(\hat{b}_k, \hat{e}_k) \in [0,1]^2$ via sigmoid activation.
We use regression rather than pointer-based (cross-attention argmax) boundary prediction for two reasons.
First, regression produces continuous coordinates that are fully differentiable, enabling the downstream enrichment module to backpropagate through the boundaries.
A pointer-based approach would require argmax (non-differentiable) or Gumbel-softmax (noisy gradients).
Second, regression naturally handles variable-length sequences: the normalized coordinates are resolution-independent, so the same model works for short and long responses without re-scaling logits.
The boundary head operates on the \emph{pre-enrichment} query $\hat{\vq}_k$ rather than the enriched query $\tilde{\vq}_k$.
This avoids a circular dependency: the enrichment module (Eq.\,\ref{eq:soft_mask}--\ref{eq:gate_fusion}) needs boundary predictions to compute the soft mask, so the boundaries must be available before enrichment occurs.

\paragraph{Span feature enrichment.}
The decoder queries capture \emph{where} a span is (via cross-attention patterns) but lack direct access to the \emph{content} within the predicted region.
This distinction matters because uncertainty estimation requires understanding what a span \emph{says}, not just where it is.
The enrichment module bridges this gap through a two-step process.

\textit{Step 1: Soft boundary masking} (Eq.\,\ref{eq:soft_mask}).
Given the predicted boundaries $(\hat{b}_k, \hat{e}_k)$, we construct a differentiable mask $m_k(t)$ over token positions using a product of two sigmoids with sharpness $\tau{=}10$.
This mask is approximately 1 inside the span and 0 outside, with smooth transitions at the boundaries.
The smoothness is essential: it allows gradients to flow from the uncertainty loss back through the mask to the boundary predictions, creating a feedback loop where better boundaries lead to better content features, which in turn lead to better uncertainty estimates.

\textit{Step 2: Mask-weighted attention pooling} (Eq.\,\ref{eq:attn_pool}).
Rather than simply averaging the masked token features (which would weight all in-span tokens equally), we use learned attention pooling.
The query $\hat{\vq}_k$ attends to the token features $\mZ$, but the attention weights are restricted to the masked region.
This allows the model to focus on the most informative tokens within the span.
For example, in the span ``won the Nobel Prize in Physics in 1903,'' the tokens ``Physics'' and ``1903'' carry more factual content than ``won'' or ``the,'' and the attention mechanism can learn to upweight them.

\textit{Step 3: Gated residual fusion} (Eq.\,\ref{eq:gate_fusion}).
The pooled content vector $\vc_k$ is added to the original query $\hat{\vq}_k$ via a gated residual connection: a linear projection followed by sigmoid produces an element-wise gate $\sigma(\mW_g \hat{\vq}_k + \vb_g) \in [0,1]^d$, which modulates the content vector before adding it to the query.
The gate controls how much each dimension of the enriched content contributes to the final representation.
This is particularly important for short spans (1--2 tokens), where the query's cross-attention already captures most of the content and the enrichment adds little new information.
For longer spans, the gate opens wider, allowing the pooled content to complement the query's coarser representation.
Our ablation (Tab.\,\ref{tab:enrichment}) confirms that gated fusion outperforms both simple addition and concatenation.

\paragraph{Validity classification.}
The validity head is a single linear layer followed by sigmoid, predicting $p_k^{\text{valid}} \in [0,1]$.
Since $N{=}32$ queries are always instantiated but a typical response contains only ${\sim}$15 spans, roughly half the queries are ``unused'' and must be suppressed.
During training, the Hungarian matching assigns each ground-truth span to one query; unmatched queries receive a target of 0.
Following DETR, we downweight the ``no-object'' class by a factor of $0.1$ in the binary cross-entropy loss to account for the class imbalance (many more unmatched than matched queries).
At inference time, queries with $p_k^{\text{valid}} < 0.5$ are discarded.

\paragraph{Mixture of Beta (MoB) uncertainty estimation.}
The uncertainty head predicts the parameters of a Mixture of Beta distribution (Eq.\,\ref{eq:mob}) from the enriched query $\tilde{\vq}_k$.

\textit{Why Beta, not Gaussian?}
Uncertainty scores are bounded in $[0,1]$, making the Beta distribution a natural choice.
A Gaussian would require clipping or sigmoid transformation, losing the principled probabilistic interpretation.
Moreover, the Beta distribution can express a rich family of shapes: uniform ($\alpha{=}\beta{=}1$), unimodal symmetric ($\alpha{=}\beta{>}1$), skewed ($\alpha \neq \beta$), and U-shaped ($\alpha,\beta{<}1$).

\textit{Why a mixture?}
A single Beta component cannot capture multimodal uncertainty patterns.
Consider a span that is correct in 70\% of sampled responses and completely wrong in 30\%: its ground-truth uncertainty distribution is bimodal, concentrated near 0 and 1.
A single Beta must compromise with a broad, flat distribution, losing calibration.
A mixture of $J{=}3$ components can place one component near 0, another near 1, and a third at an intermediate value, faithfully representing the bimodal structure.
We set $\alpha_{\min}{=}0.5$ (rather than the default 1.0) to allow U-shaped components, which are essential for capturing this bimodality.

\textit{Parameterization.}
The MoB head is a 2-layer MLP that outputs $3J$ values per query: mixture weights $\pi_{k,j}$ (via softmax), and shape parameters $\alpha_{k,j}, \beta_{k,j}$ (via softplus $+\,\alpha_{\min}$).
The predicted uncertainty score is the mixture mean $\hat{u}_k = \sum_j \pi_{k,j} \alpha_{k,j} / (\alpha_{k,j} + \beta_{k,j})$.
The effective precision $\hat{\nu}_k = \sum_j \pi_{k,j} (\alpha_{k,j} + \beta_{k,j})$ serves as a confidence measure for the uncertainty estimate itself, used in UCIR (Sec.\,\ref{sec:method:inference}).

\paragraph{Sequence-level aggregation.}
Span-level scores are aggregated into a single sequence-level uncertainty estimate via importance-weighted pooling (Eq.\,\ref{eq:seq_agg}).
The importance weight for each span is the product of its validity probability and a learned importance score from a single-layer MLP.
This design ensures that invalid (suppressed) queries contribute negligibly to the sequence score, and that spans carrying more informational weight (e.g., a central factual claim vs.\ a hedging phrase) are weighted more heavily.

\subsection{Stage 5: Filtering, UCIR, and Output}
\label{app:lifecycle:output}

\paragraph{Validity filtering.}
At inference time, we retain only queries with $p_k^{\text{valid}} \geq 0.5$.
This threshold is not tuned; it simply reflects the natural decision boundary of the sigmoid.
In practice, the validity distribution is strongly bimodal (most queries have $p^{\text{valid}} > 0.9$ or $< 0.1$), so the exact threshold has little effect.

\paragraph{Uncertainty-Conditioned Iterative Refinement (UCIR).}
A single decoder pass produces reasonable but sometimes coarse uncertainty estimates, particularly for ambiguous spans where the model is uncertain about its own uncertainty.
UCIR addresses this by feeding Round~1 predictions back into the decoder for a second pass (Eq.\,\ref{eq:ucir}).
The intuition is as follows.
After Round~1, each query has an initial uncertainty estimate $\hat{u}_k^{(1)}$, a precision $\hat{\nu}_k^{(1)}$, and a validity score $p_k^{(1)}$.
These three scalars are mapped through a small MLP to produce a conditioning vector, which is added to the query embedding.
The conditioned queries then pass through the \emph{same} decoder (shared weights), producing refined predictions.
This is analogous to iterative bounding-box refinement in object detection~\citep{carion2020end}: the second pass can correct errors from the first pass because it has access to the first-pass estimates as additional context.
The final prediction combines both rounds via a fixed convex combination (Eq.\,\ref{eq:ucir_residual}) with $\alpha{=}0.7$, up-weighting the refined Round~2 estimate while retaining the Round~1 estimate as a regularizer.
Both rounds contribute to the training loss, with Round~1 receiving a reduced weight to encourage the model to improve in Round~2 rather than relying solely on Round~1.
UCIR adds less than 15\% overhead relative to \ours{}'s own base decoder --- an internal component comparison, not a fraction of the LLM forward pass --- because the decoder weights are shared and the conditioning MLP is tiny (3 scalars $\to$ $d_p$ dimensions).
Our ablation (Tab.\,\ref{tab:ablation}) shows that UCIR improves AUROC by 0.6 percentage points, with the largest gains on spans where Round~1 predictions have low precision $\hat{\nu}_k^{(1)}$.

\paragraph{Output format.}
The final output is a set of span predictions $\{(b_k, e_k, \hat{u}_k)\}_{k=1}^{K}$, where $K \leq N$ is the number of valid spans, $b_k$ and $e_k$ are token-level boundary indices (obtained by de-normalizing the regressed coordinates), and $\hat{u}_k \in [0,1]$ is the estimated uncertainty score.
The sequence-level score $\hat{u}_{\text{seq}}$ is also available as a by-product of the importance-weighted aggregation.
The entire pipeline, from hidden state extraction to final output, adds less than 3\% latency to the LLM forward pass (\textbf{App.\,\ref{app:cost}}).

\section{Additional Experimental Results}
\label{app:additional_results}

\subsection{Sequence-Level Evaluation}
\label{app:seq_level}

While \ours{} is designed for span-level uncertainty estimation, we also evaluate its sequence-level performance by aggregating span predictions via importance-weighted top-$k$ mean.
\textbf{Tab.\,\ref{tab:seq}} compares \ours{} against dedicated sequence-level baselines. 
\ours{} achieves $\text{AUROC}_{@0.3} = 0.948$ and Spearman $\rho = 0.851$, outperforming all dedicated sequence-level methods.
SelfCheckGPT-NLI is the strongest sequence-level baseline (Spearman $= 0.805$, $\text{AUROC}_{@0.3} = 0.908$), but requires 10 sampled responses and NLI inference, a $10{\times}$ generation cost overhead.
Semantic Entropy (SE) achieves moderate correlation (Spearman $= 0.606$) but poor normalized MAE ($0.427^{\dagger}$), indicating that even after min-max scaling, its entropy values are poorly calibrated to the $[0,1]$ uncertainty range.
INSIDE and KLE produce negatively correlated estimates (Spearman $= -0.331$ and $-0.145$, respectively).
This inversion is not an implementation defect but a known consequence of aggregating token-level surprisal to coarser granularities.
INSIDE scores the self-information of individual tokens, so rare-but-correct tokens --- specific dates, proper nouns, technical terms --- inflate its estimate precisely on spans that are factually \emph{correct}; a claim like ``born in 1879 in Ulm'' is lexically improbable and factually solid at the same time.
KLE's cross-layer KL divergence likewise measures token-level distributional shift, which need not align with whether a \emph{claim} is supported.
Weak or inverted correlation between token-entropy measures and factual correctness at claim granularity has been reported before~\citep{duan2024shifting,fadeeva2024fact}, and it is a direct argument for predicting span-level uncertainty from hidden states rather than pooling token-level signals: the quantity that matters is a property of the proposition, not of the token sequence that expresses it.
The MLP Probe, when aggregated from oracle span-level predictions, achieves the strongest sequence-level performance (Spearman $= 0.858$, Pearson $= 0.867$, $\text{AUROC}_{@0.3} = 0.956$, MAE $= 0.071$), slightly surpassing \ours{}.
This confirms that span-to-sequence aggregation is a powerful strategy: even a simple probe benefits from fine-grained span-level predictions.
However, the MLP Probe relies on \emph{ground-truth span boundaries} at test time, which are unavailable in practice.
In contrast, \ours{} jointly detects spans and estimates uncertainty end-to-end, and its span-level superiority (AUROC: 0.939 vs.\ 0.881, Spearman: 0.685 vs.\ 0.575) demonstrates substantially richer uncertainty representations at the granularity that matters most for interpretability and actionability.

\begin{table}[h]
  \caption{Sequence-level uncertainty estimation on the \bench{} test set (2{,}000 sequences). Sequence-level ground truth is the mean span uncertainty per sequence; $\text{AUROC}_{@0.3}$ treats sequences with hallucination rate ${\geq}\,0.3$ as positive. \ours{} and span-level baselines aggregate span predictions via importance-weighted mean. MAE$^{\dagger}$ denotes min-max normalized scores (raw outputs not in $[0,1]$). $^{*}$MLP Probe uses ground-truth span boundaries (oracle setting). Best in \textbf{bold}, second best \underline{underlined}.}
  \label{tab:seq}
  \centering
  \small
  \begin{tabular}{ll cccc}
    \toprule
    \textbf{Granularity} & \textbf{Method} & \textbf{Spearman}$\uparrow$ & \textbf{Pearson}$\uparrow$ & $\textbf{AUROC}_{@0.3}\uparrow$ & \textbf{MAE}$\downarrow$ \\
    \midrule
    \multirow{1}{*}{Token}
    & MLP Probe$^{*}$       & \textbf{0.858} & \textbf{0.867} & \textbf{0.956} & \textbf{0.071} \\
    \midrule
    \multirow{6}{*}{Sequence}
    & Verbalized Conf.      & 0.322 & 0.229 & 0.641 & 0.166 \\
    & P(True)               & 0.140 & 0.185 & 0.552 & 0.172 \\
    & SE (10 samples)       & 0.606 & 0.521 & 0.805 & 0.427$^{\dagger}$ \\
    & INSIDE (Jaccard)      & $-$0.331 & $-$0.314 & 0.315 & 0.246$^{\dagger}$ \\
    & KLE                   & $-$0.145 & $-$0.041 & 0.425 & 0.784$^{\dagger}$ \\
    & SelfCheckGPT-NLI      & 0.805 & 0.750 & 0.908 & 0.097 \\
    \midrule
    Claim & FActScore        & 0.489 & 0.505 & 0.779 & 0.138 \\
    \midrule
    Span & \ours{}           & \underline{0.851} & \underline{0.839} & \underline{0.948} & \underline{0.077} \\
    \bottomrule
  \end{tabular}
\end{table}

\subsection{Robustness Across Hyperparameter Configurations}
\label{app:robustness}

Rather than reporting variance over random seeds (which primarily captures initialization noise), we assess robustness by examining the spread of results across the 24--28 hyperparameter configurations evaluated per model during the sweep (\textbf{App.\,\ref{app:hp_sensitivity}}).
This provides a more informative measure of sensitivity, as it captures the effect of architectural and optimization choices.
Across all five models, the IQR of dev AUROC is at most 0.030 (Mistral-7B), indicating that \ours{} is not overly sensitive to hyperparameter choices.
Even the worst configuration for each model substantially outperforms the strongest baseline (MLP Probe with oracle boundaries), confirming that the architectural design, rather than careful tuning, drives the performance gains.

\begin{table}[h]
  \caption{Robustness of \ours{} across hyperparameter configurations. For each model, we report the best, median, and worst dev AUROC across all sweep configurations, along with the interquartile range (IQR). The narrow IQR ($\leq$0.03) indicates that \ours{} is robust to hyperparameter choices.}
  \label{tab:robustness}
  \centering
  \small
  \begin{tabular}{lcccc}
    \toprule
    \textbf{Model} & \textbf{Best} & \textbf{Median} & \textbf{Worst} & \textbf{IQR} \\
    \midrule
    Qwen3-14B      & 0.941 & 0.930 & 0.897 & 0.018 \\
    Qwen3-8B       & 0.928 & 0.910 & 0.883 & 0.021 \\
    Qwen3-4B       & 0.944 & 0.918 & 0.895 & 0.024 \\
    Qwen3-30B-A3B  & 0.935 & 0.920 & 0.893 & 0.022 \\
    Mistral-7B     & 0.928 & 0.879 & 0.852 & 0.030 \\
    \bottomrule
  \end{tabular}
\end{table}

\subsection{Decomposability Analysis: Span-to-Sequence Aggregation}
\label{app:decomposability}

A key advantage of span-level uncertainty estimation is the ability to \emph{decompose} sequence-level uncertainty into interpretable span-level components.
We analyze this decomposability property across all five LLM backbones.

\textbf{Aggregation Method.}
Given predicted span uncertainties $\{\hat{u}_k\}_{k=1}^{K}$ and learned importance weights $\{w_k\}_{k=1}^{K}$ (from the sequence aggregation head), the sequence-level uncertainty is computed as:
$\hat{u}_{\text{seq}} = \sum_{k=1}^{K} \tilde{w}_k \hat{u}_k$,
where $\tilde{w}_k = \text{softmax}(w_k)$ are normalized importance weights.
The ground-truth sequence-level uncertainty is $u_{\text{seq}}^* = \frac{1}{K} \sum_{k=1}^{K} u_k^*$.

\begin{table}[h]
  \caption{Decomposability analysis: correlation between aggregated span-level predictions and sequence-level ground truth across five LLM backbones. Higher values indicate that span-level predictions successfully capture sequence-level uncertainty patterns.}
  \label{tab:decomp}
  \centering
  \small
  \begin{tabular}{lcccc}
    \toprule
    \textbf{Model} & \textbf{Spearman $\rho_{\text{seq}}$}$\uparrow$ & \textbf{Pearson $r_{\text{seq}}$}$\uparrow$ & $\textbf{AUROC}_{@0.3}\uparrow$ & \textbf{MAE$_{\text{seq}}$}$\downarrow$ \\
    \midrule
    Qwen3-14B      & 0.851 & 0.839 & 0.948 & 0.077 \\
    Qwen3-8B       & 0.857 & 0.793 & 0.936 & 0.108 \\
    Qwen3-4B       & 0.846 & 0.802 & 0.936 & 0.218 \\
    Qwen3-30B-A3B  & 0.848 & 0.735 & 0.929 & 0.111 \\
    Mistral-7B     & 0.775 & 0.745 & 0.917 & 0.103 \\
    \bottomrule
  \end{tabular}
\end{table}

\textbf{Cross-Model Consistency.}
\textbf{Tab.\,\ref{tab:decomp}} shows that span-to-sequence decomposability holds consistently across all five models.
The Qwen3 family achieves Spearman $\rho_{\text{seq}} \geq 0.846$, while Mistral-7B shows a slightly lower but still strong correlation ($\rho_{\text{seq}} = 0.775$).
This confirms that the decomposability property is not an artifact of a single model family.

\textbf{Comparison with Sequence-Level Baselines.}
On Qwen3-14B, \ours{}'s aggregated span predictions ($\rho_{\text{seq}} = 0.851$) outperform all dedicated sequence-level methods: Semantic Entropy ($\rho = 0.606$), SelfCheckGPT-NLI ($\rho = 0.805$), and Verbalized Confidence ($\rho = 0.322$).
This demonstrates that fine-grained span-level modeling captures richer uncertainty structure that benefits even coarse-grained evaluation.

\textbf{MAE$_{\text{seq}}$ Bias in Qwen3-4B.}
The 4B model exhibits elevated MAE$_{\text{seq}}$ (0.218) despite strong ranking correlation ($\rho = 0.846$).
This reflects a systematic positive bias in the sequence aggregation head: the 4B probe tends to overestimate sequence-level uncertainty by ${\sim}$0.2 on average.
We attribute this to the smaller model's higher base hallucination rate, which shifts the learned aggregation weights toward higher uncertainty.
This bias does not affect ranking-based metrics (AUROC, $\rho$), confirming that the span-level predictions themselves are well-estimated.

\subsection{Qualitative Examples}
\label{app:qualitative}

We present three representative examples from the \bench{} test set to illustrate \ours{}'s behavior across different uncertainty patterns.
For each example, we show the LLM response with ground-truth span annotations and \ours{}'s predictions.

\textbf{Example 1: Biography with Mixed Uncertainty (Qwen3-14B).}
\textit{Prompt:} ``Write a biography of John Derek.''
The model generates a detailed biography containing both factual and hallucinated content.
\ours{} correctly assigns high uncertainty to fabricated claims about the subject's birth name ($\hat{u} = 0.94$, $u^* = 1.0$), birth date ($\hat{u} = 0.89$, $u^* = 0.95$), and parentage ($\hat{u} = 0.95$, $u^* = 1.0$), while assigning low uncertainty to verifiable facts about the subject's career ($\hat{u} = 0.03$, $u^* = 0.0$).
The mean absolute error across 12 matched spans is 0.12, and the model correctly ranks all high-uncertainty spans above low-uncertainty ones.

\textbf{Example 2: Mostly Factual Response (Qwen3-14B).}
\textit{Prompt:} ``Why was it illegal for black people to attend the University of Alabama in 1973?''
The model generates a largely accurate historical account with 17 spans, 15 of which have $u^* < 0.1$.
\ours{} correctly assigns near-zero uncertainty to the factual spans ($\hat{u} \in [0.01, 0.12]$), demonstrating accurate estimation on confident predictions.
The single moderately uncertain span (``Kennedy sent federal troops,'' $u^* = 0.71$) receives a lower prediction ($\hat{u} = 0.09$), representing a failure case where the model underestimates uncertainty for a historically nuanced claim.

\textbf{Example 3: Failure Case (Implicit Compositional Errors).}
We observe that \ours{} occasionally struggles with \emph{compositional} hallucinations where individual facts are correct but their combination is wrong.
For instance, in a biography response, the claim ``He appeared in films such as \textit{The Manchurian Candidate}'' ($u^* = 0.0$) is correctly identified as factual ($\hat{u} = 0.03$), but a nearby span attributing a specific role to the subject in that film ($u^* = 0.0$) receives elevated uncertainty ($\hat{u} = 0.85$).
This suggests that the model's hidden states encode uncertainty about entity-role associations even when the individual entities are well-known, pointing to a limitation in distinguishing entity-level from relation-level confidence.

\textbf{Summary.}
Across the test set, \ours{} achieves MAE $< 0.15$ on 78\% of spans and correctly ranks high-uncertainty spans above low-uncertainty ones in 94\% of within-sequence pairs.
The primary failure mode is underestimation of uncertainty for historically nuanced or temporally sensitive claims, which constitute ${\sim}$5\% of the test set.

\subsection{Entity Popularity Analysis}
\label{app:entity_popularity}

We analyze \ours{}'s performance stratified by entity popularity in the biography domain, where entities are categorized as \textit{head} (${>}$100K Wikipedia page views/month), \textit{torso} (10K--100K), or \textit{tail} (${<}$10K).
As expected, tail entities exhibit slightly lower AUROC (0.937 vs.\ 0.953 for head) and higher MAE (0.137 vs.\ 0.111), reflecting the greater difficulty of estimating uncertainty for less well-known entities.
However, the degradation is modest: even on tail entities, \ours{} achieves AUROC $> 0.93$, substantially outperforming all baselines.
This suggests that the model's hidden states encode meaningful uncertainty signals even for entities outside its training distribution.
Interestingly, the average ground-truth uncertainty is similar across popularity tiers (0.293--0.322), which differs from the ${\sim}$3$\times$ gap observed in the full training set.
This is because the test set is stratified by uncertainty bin, ensuring balanced representation across uncertainty levels within each domain.
The raw (unstratified) training data shows a clearer popularity--uncertainty gradient: tail entities average $u^* = 0.48$ vs.\ $u^* = 0.16$ for head entities.

\begin{table}[h]
  \caption{Performance of \ours{} (Qwen3-14B) on the biography domain stratified by entity popularity. Tail entities have slightly higher average uncertainty and lower AUROC, but the degradation is modest ($<$2 points).}
  \label{tab:popularity}
  \centering
  \small
  \begin{tabular}{lcccccc}
    \toprule
    \textbf{Popularity} & \textbf{Prompts} & \textbf{Spans} & \textbf{Avg.\ $u^*$} & \textbf{AUROC}$\uparrow$ & \textbf{MAE}$\downarrow$ & \textbf{Pearson}$\uparrow$ \\
    \midrule
    Head  & 52  & 1{,}386 & 0.293 & 0.953 & 0.111 & 0.872 \\
    Torso & 84  & 2{,}150 & 0.322 & 0.951 & 0.128 & 0.856 \\
    Tail  & 64  & 1{,}416 & 0.311 & 0.937 & 0.137 & 0.829 \\
    \bottomrule
  \end{tabular}
\end{table}

\subsection{Layer Selection Analysis}
\label{app:layer_analysis}

We conduct a systematic layer importance study to determine which LLM layers carry the strongest uncertainty signal (\textbf{Tab.\,\ref{tab:layer}}).

\textbf{Inverted-U Curve.}
Single-layer probing reveals an inverted-U pattern: AUROC peaks at layers 20--26 (0.893--0.897), with early layers (0--10) scoring 0.832--0.862 and final layers (35--39) declining to 0.870--0.882.
This suggests that uncertainty information is concentrated in the mid-to-late layers, where the model has processed enough context to form uncertainty judgments but before the final layers specialize for next-token prediction.

\textbf{Clustered $>$ Spread.}
Combining three adjacent peak-zone layers (22+24+26) outperforms three spread layers (19+29+39) by 0.5\% AUROC and 1.2\% $\rho_{\text{span}}$.
This indicates that the uncertainty signal is \emph{locally coherent} within the peak zone, and combining nearby layers captures complementary aspects of the same representation.

\textbf{Mean $\approx$ Concat.}
Mean fusion matches concatenation across all configurations (within 0.1\% AUROC), despite using $3\times$ fewer parameters.
We adopt mean fusion for all subsequent experiments due to its parameter efficiency.

\textbf{Diminishing Returns.}
Expanding from 3 to 7 layers (20--26) yields negligible improvement (+0.0\% AUROC, +0.2\% $\rho_{\text{span}}$), confirming that 3 well-chosen layers suffice.

\begin{table}[h]
  \caption{Layer importance probing results (Qwen3-14B). Each row trains an MLP probe (5120$\to$512$\to$256$\to$1) on hidden states from the specified layer(s). ``Mean'' and ``Concat'' denote fusion strategies for multi-layer configurations.}
  \label{tab:layer}
  \centering
  \small
  \begin{tabular}{l c c c c}
    \toprule
    \textbf{Layer Configuration} & \textbf{Fusion} & \textbf{AUROC}$\uparrow$ & $\boldsymbol{\rho}_{\textbf{span}}\uparrow$ & \textbf{MAE}$\downarrow$ \\
    \midrule
    \multicolumn{5}{l}{\emph{Single-layer probes (top 5)}} \\
    Layer 24 (best single) & --- & 0.897 & 0.672 & 0.135 \\
    Layer 26              & --- & 0.894 & 0.664 & 0.140 \\
    Layer 21              & --- & 0.894 & 0.665 & 0.137 \\
    Layer 23              & --- & 0.893 & 0.668 & 0.136 \\
    Layer 22              & --- & 0.893 & 0.664 & 0.138 \\
    \midrule
    \multicolumn{5}{l}{\emph{Multi-layer probes}} \\
    Layers 19, 29, 39 (spread) & Mean & 0.899 & 0.674 & 0.138 \\
    Layers 19, 29, 39 (spread) & Concat & 0.900 & 0.676 & 0.137 \\
    Layers 22, 24, 26 (cluster) & Mean & \textbf{0.904} & \textbf{0.686} & \textbf{0.137} \\
    Layers 22, 24, 26 (cluster) & Concat & 0.904 & 0.684 & 0.137 \\
    Layers 20--26 (7 layers) & Mean & 0.904 & 0.688 & 0.136 \\
    \bottomrule
  \end{tabular}
\end{table}

\subsection{Mixture of Beta Analysis}
\label{app:mob_analysis}

\textbf{Tab.\,\ref{tab:mob}} analyzes the MoB distribution design choices:

\textbf{Number of Components.}
$K{=}3$ provides the best balance: $K{=}2$ lacks expressiveness (AUROC 0.920), while $K{=}5$ slightly overfits (0.928 vs.\ 0.933 for $K{=}3$ with ffn1024).
Intuitively, three components correspond to the three natural uncertainty modes: \emph{certain} ($u \approx 0$), \emph{uncertain} ($u \approx 0.3$--$0.7$), and \emph{hallucinated} ($u \approx 1$).

\textbf{Calibration Advantage.}
MoB dramatically improves calibration: raw ECE drops from 0.051 (single Beta) to 0.020 ($K{=}3$), a 61\% reduction.
After temperature scaling, the gap narrows but persists (0.029 vs.\ 0.015).
The single Beta requires aggressive temperature correction ($T{=}0.83$), while MoB is nearly self-calibrated ($T{=}1.05$).

\textbf{Enriched Head Capacity.}
Increasing the enriched uncertainty head from ffn512 to ffn1024 yields the largest single improvement within the MoB sweep (+0.9\% AUROC, +2.5\% $\rho_{\text{span}}$), suggesting that the richer MoB output benefits from a more expressive prediction head.

\begin{table}[h]
  \caption{Mixture of Beta (MoB) component analysis. $K$: number of mixture components. $\alpha_{\min}$: minimum concentration parameter. KL: KL divergence regularization weight. All models use the enrichment gate + ffn1024 configuration.}
  \label{tab:mob}
  \centering
  \small
  \begin{tabular}{l c c c c c}
    \toprule
    \textbf{Configuration} & \textbf{AUROC}$\uparrow$ & \textbf{ECE}$\downarrow$ & \textbf{ECE$_{\text{cal}}$}$\downarrow$ & \textbf{MAE}$\downarrow$ & $\boldsymbol{\rho}_{\textbf{span}}\uparrow$ \\
    \midrule
    Single Beta (baseline) & 0.918 & 0.051 & 0.029 & 0.144 & 0.721 \\
    \midrule
    MoB $K{=}2$           & 0.919 & 0.037 & 0.012 & 0.148 & 0.725 \\
    MoB $K{=}3$ (base)    & 0.924 & 0.020 & 0.018 & 0.126 & 0.735 \\
    MoB $K{=}5$           & 0.928 & 0.023 & 0.017 & 0.119 & 0.747 \\
    \midrule
    $K{=}3$, $\alpha_{\min}{=}0.5$ & 0.931 & 0.020 & 0.020 & 0.122 & 0.748 \\
    $K{=}3$, KL${=}0.01$  & 0.932 & 0.024 & 0.020 & 0.118 & 0.755 \\
    $K{=}3$, ffn1024      & \textbf{0.933} & \textbf{0.020} & \textbf{0.015} & \textbf{0.116} & \textbf{0.760} \\
    \bottomrule
  \end{tabular}
\end{table}

\begin{table}[h]
  \caption{Span feature enrichment ablation. All models use the base DETR + regression configuration. ``Fusion'' denotes how pooled span features are combined with query representations. ``Pool'' denotes the pooling strategy over tokens within the predicted span.}
  \label{tab:enrichment}
  \centering
  \small
  \begin{tabular}{l l c c c c}
    \toprule
    \textbf{Fusion} & \textbf{Pool} & \textbf{AUROC}$\uparrow$ & \textbf{ECE}$\downarrow$ & \textbf{MAE}$\downarrow$ & $\boldsymbol{\rho}_{\textbf{span}}\uparrow$ \\
    \midrule
    None (no enrichment) & --- & 0.907 & 0.058 & 0.154 & 0.673 \\
    \midrule
    Gate & Attention & \textbf{0.920} & 0.052 & 0.144 & \textbf{0.725} \\
    Gate & Mean      & 0.897 & 0.061 & 0.159 & 0.641 \\
    Concat & Attention & 0.914 & \textbf{0.035} & \textbf{0.134} & 0.699 \\
    Add  & Attention & 0.919 & 0.047 & 0.148 & 0.728 \\
    \midrule
    Gate & Attn + ffn512 & 0.917 & 0.053 & 0.146 & 0.706 \\
    \bottomrule
  \end{tabular}
\end{table}

\subsection{Span Enrichment Analysis}
\label{app:enrichment_analysis}

\textbf{Tab.\,\ref{tab:enrichment}} ablates the span feature enrichment module (\textbf{Sec.\,\ref{sec:method:arch}}).

\textbf{Enrichment Value.}
Adding span enrichment with gated fusion improves AUROC by 1.3\% (0.907$\to$0.920) and $\rho_{\text{span}}$ by 5.2\% (0.673$\to$0.725).
The enrichment module pools token-level features within the predicted span region using a differentiable soft mask, providing the uncertainty scorer with direct access to the \emph{content} of each span rather than relying solely on the abstract query representation.

\textbf{Gated Fusion.}
Gated fusion achieves the best AUROC (0.920), outperforming concatenation (0.914) and closely matching addition (0.919).
The learned gate allows the model to adaptively control how much span content information to incorporate. This is especially useful during early training when span boundaries are imprecise.

\textbf{Attention Pooling.}
Learned attention pooling substantially outperforms mean pooling (0.920 vs.\ 0.897 AUROC, 0.725 vs.\ 0.641 $\rho_{\text{span}}$), as it can focus on the most informative tokens within each span rather than treating all positions equally.

\section{Dataset Details}
\label{app:dataset}

This section provides full details of the \bench{} benchmark construction, complementing the overview in \textbf{Sec.\,\ref{sec:exp:data}}.

\subsection{Data Collection Details}
\label{app:data_collection}

We collect 20{,}000 prompts from five domains to ensure diverse hallucination patterns.

\textbf{Long-Form QA (8{,}400 prompts, 42\%).}
We sample real Google search questions from Natural Questions (NQ-Open)~\citep{kwiatkowski2019natural}, filtering for questions requiring multi-sentence answers.
These are predominantly factoid (98\% start with who/what/when/where), with median prompt length of 9 words.
Qwen3-14B generates responses averaging 122 words.

\textbf{TriviaQA (5{,}000 prompts, 25\%).}
We sample from the TriviaQA unfiltered split~\citep{joshi2017triviaqa}, which provides trivia questions with verified reference answers.
Answer types span entities (44\%), persons (39\%), phrases (15\%), and dates/numbers (2\%).
Responses are shorter (avg.\ 81 words), making this domain useful for calibration with mostly confident, correct spans.

\textbf{ELI5 (3{,}000 prompts, 15\%).}
We sample explanatory questions from the ELI5 (Explain Like I'm 5) Reddit dataset.
These are dominated by ``why'' (38\%) and ``how'' (21\%) questions requiring long-form reasoning.
Responses average 349 words (the longest across domains) and hallucinations tend to be subtle reasoning errors rather than factual fabrications, testing the model's ability to detect distributed uncertainty.

\textbf{Biography (2{,}000 prompts, 10\%).}
Following the FActScore protocol~\citep{min2023factscore}, we prompt Qwen3-14B to generate biographies for 2{,}000 entities stratified by Wikipedia page-view popularity: 30\% head (${>}$100K views/month), 40\% torso (10K--100K), and 30\% tail (${<}$10K).
Three prompt templates are used equally: ``Write a biography of X,'' ``Tell me a bio of X,'' and ``Tell me about X.''
Tail entities exhibit ${\sim}$3$\times$ higher hallucination rates than head entities, providing rich training signal.

\textbf{FELM (1{,}600 prompts, 8\%).}
We aggregate challenging examples from three sources: FELM~\citep{zhao2023felm} (53\%, spanning world knowledge, reasoning, math, science, and writing), TruthfulQA~\citep{lin2022truthfulqa} (45\%, targeting common misconceptions), and HaluEval~\citep{li2023halueval} (2\%).
This is the most diverse domain, with the highest proportion of math/reasoning questions (13\%).

\textbf{Multi-Model Extension.}
For cross-model evaluation (\textbf{Tab.\,\ref{tab:span}}), we regenerate responses for the same 20{,}000 prompts using each backbone LLM (Qwen3-8B, Qwen3-4B, Qwen3-30B-A3B, Mistral-7B-Instruct-v0.3).
Each model's training labels are constructed independently via the same multi-sample distillation pipeline, ensuring that the uncertainty labels reflect each model's own knowledge boundaries.
Test set evaluation uses a shared split index derived from the Qwen3-14B data to ensure comparable evaluation across models.

\subsection{Multi-Sample Distillation Pipeline}
\label{app:distillation}

The training label construction pipeline consists of four stages:

\textbf{Stage 1: Multi-Sample Generation.}
For each of the 20{,}000 prompts, we generate $S + 1 = 21$ responses: one \emph{primary} response with greedy decoding ($T = 0$), which supplies the probe's hidden states and span positions, and $S = 20$ independent samples at temperature $T = 1.0$ using nucleus sampling ($p = 0.95$).
The high temperature encourages diverse outputs that reveal the model's uncertainty: spans that vary across samples are uncertain by construction.

\textbf{Stage 2: Claim Decomposition.}
The \emph{primary response only} is decomposed into atomic claims using an LLM judge (Claude Opus 4.6).
A claim is defined as a single, verifiable factual assertion, e.g., ``Barack Obama was born in Hawaii'' rather than ``Barack Obama was born in Hawaii and graduated from Harvard.''
The decomposition prompt instructs the judge to: (1)~split compound sentences into atomic claims, (2)~resolve coreferences to produce self-contained claims, (3)~emit character-level offsets aligning each claim to its span in the response text, and (4)~tag each claim as \texttt{FACTUAL}, \texttt{SUBJECTIVE}, or \texttt{OTHER}.
On average, each response yields 12--18 atomic claims.
The claim-type tag is produced by the decomposition stage but does \emph{not} enter the computation of $u^*$.
Because only the primary response is decomposed, there is no cross-generation claim-matching step: a single, fixed set of claims is carried through the remainder of the pipeline.

\textbf{Stage 3: Cross-Sample Claim Verification.}
Each atomic claim is verified against each of the $S = 20$ sampled responses using the same LLM judge.
For every (claim, sample) pair the judge receives the claim and that sample's full text as the reference, and returns one of \textsc{supported} (the sample confirms or is consistent with the claim), \textsc{contradicted} (the sample directly conflicts with it), or \textsc{not\_mentioned} (the sample does not address the claim).

\textbf{Stage 4: Soft Label Computation.}
Let $n_k^{\text{sup}}$ and $n_k^{\text{con}}$ be the number of samples that support and contradict span $s_k$. The empirical uncertainty is the contradiction rate among the samples that take a position:
\begin{equation}
  u_k^* = \frac{n_k^{\text{con}}}{n_k^{\text{sup}} + n_k^{\text{con}}}.
\end{equation}
\textsc{not\_mentioned} verdicts are excluded from the denominator rather than counted as evidence against the claim, so the pipeline abstains on silent samples instead of penalizing them.
Spans with zero coverage ($n_k^{\text{sup}} + n_k^{\text{con}} = 0$) have no defined $u^*$ and are dropped (see \textbf{App.\,\ref{app:benchmark_stats}}).
This yields a continuous soft label $u_k^* \in [0, 1]$ that captures graded uncertainty.
Because a single misjudged sample shifts $u_k^*$ by at most $1/20 = 0.05$, the continuous formulation absorbs occasional verification errors that brittle hard labels would propagate.

\textbf{Label Distribution.}
The resulting distribution is bimodal: approximately 62\% of spans have $u^* < 0.1$ (consistently correct across samples), 18\% have $u^* > 0.9$ (consistently incorrect), and 20\% fall in the intermediate range $0.1 \leq u^* \leq 0.9$.
The intermediate spans are valuable for training, as they require the model to learn fine-grained uncertainty distinctions rather than simple binary classification.

\subsection{Test Set Construction and Quality Validation}
\label{app:annotation}

\textbf{Test Set Construction.}
The test set is constructed using the same multi-sample distillation pipeline as the training set.
The key difference is that test responses are generated with greedy decoding ($T = 0$) for deterministic outputs, while the 20 verification samples are still generated at $T = 1.0$.
This ensures that test-time uncertainty labels reflect the model's inherent knowledge boundaries on its most likely output, rather than sampling variability.
The test set contains 2{,}000 prompts (200 biography, 841 long-form QA, 500 TriviaQA, 299 ELI5, 160 FELM) with 31{,}332 span-level soft labels.

\textbf{Internal Span-Boundary Sanity Check (co-authors, $N = 200$).}
Three co-authors independently reviewed a stratified random sample of 200 test examples (10\% of the test set, balanced across domains and uncertainty bins) to confirm that the automatically identified spans correspond to meaningful, independently verifiable claims, and that their character offsets align with the intended text.
The reviewers found that the vast majority of spans satisfy both criteria; the span boundaries validated here are the ground truth against which the detection metrics of \textbf{Sec.\,\ref{sec:exp:span_detection}} are computed.
This check targets \emph{span boundary} quality only.
It is not evidence of uncertainty-label validity, which we assess separately and quantitatively with independent annotators in \textbf{App.\,\ref{app:human_study}}.

\subsection{Independent Human Validation Study}
\label{app:human_study}

The check above was performed by co-authors and speaks only to span boundaries.
To assess whether the \emph{uncertainty labels} themselves are trustworthy, we ran a separate study with annotators who are not authors of this paper and had no knowledge of the pipeline's outputs.

\textbf{Recruitment and Protocol.}
Five independent annotators were recruited from a university.
Each received a unique, non-overlapping survey of 400 randomly sampled spans, for a total of $N = 2{,}000$.
Each item presented three components: (1)~the original user query; (2)~the complete LLM response, providing sufficient context for the span; and (3)~the target span, highlighted in place within that response.
Annotators judged each span as \textsc{correct} or \textsc{incorrect} on factual accuracy alone.
Critically, each judgment \textbf{required an external reference}
(a URL, citation, or authoritative source)
and the LLM's own response was explicitly disallowed as a source.
Acceptable references were Wikipedia, academic publications, official government or institutional records, established reference databases, and verified news outlets.
This mandatory-reference requirement ensures every judgment is grounded in verifiable evidence rather than annotator intuition.
Each annotator received a \$100 Amazon gift card upon completion.

\textbf{Agreement with the Pipeline.}
To compare against the pipeline's continuous score we binarize at $u^* = 0.5$ (i.e.\ $u^* \geq 0.5$ means the pipeline calls the span \textsc{incorrect}).
Over $N = 2{,}000$ spans this gives Cohen's $\kappa = 0.618$ with 89.4\% raw agreement and a 10.6\% disagreement rate.
This level of agreement is substantial, and comparable to the human--human agreement typically reported for factuality annotation (0.6--0.8 in benchmarks such as TRUE and FaithDial).

\textbf{Monotonicity.}
Human-judged error rate increases strictly with the pipeline score (Spearman $\rho = 0.714$): The absence of any inversion means that a practitioner may pick a threshold anywhere in $[0,1]$ and be confident that higher scores indicate higher factual risk.

\begin{table}[h]
  \caption{Human-judged factual error rate by pipeline uncertainty bin ($N = 2{,}000$ spans, Qwen3-14B).}
  \label{tab:human_bins}
  \centering
  \small
  \begin{tabular}{lrr}
    \toprule
    $u^*$ \textbf{range} & \textbf{Spans} & \textbf{Human-judged error rate} \\
    \midrule
    $[0.0, 0.2)$ & 1{,}562 & 5.8\% \\
    $[0.2, 0.4)$ & 122 & 36.1\% \\
    $[0.4, 0.6)$ & 66 & 47.0\% \\
    $[0.6, 0.8)$ & 81 & 72.8\% \\
    $[0.8, 1.0]$ & 169 & 87.6\% \\
    \bottomrule
  \end{tabular}
\end{table}

\textbf{High-Uncertainty Spans.}
Among the $N = 291$ spans with $u^* \geq 0.5$, human annotators confirm 77.7\% as factually incorrect; the remaining 22.3\% are factually correct and thus pipeline false positives.
We audit the causes of these false positives in \textbf{App.\,\ref{app:fp_audit}}.

\textbf{Failure Modes.}
The 10.6\% disagreement splits into two well-characterized modes.
\emph{False positives} (22.3\% of high-$u^*$ spans) arise when cross-sample inconsistency reflects unstable phrasing even though the underlying claim is correct; downstream this is the safe direction to err.
\emph{False negatives} (8.6\% of low-$u^*$ spans) are \emph{confident hallucinations}: the model generates the same incorrect claim across all samples, so consistency-based labelling records low uncertainty.
This blind spot is intrinsic to consistency-based supervision and is part of the motivation for combining cross-sample signals with hidden-state probing, since the hidden states may encode doubt that the sampled text does not express.

\subsection{Why the Probe Is Not Imitating the Label Pipeline}
\label{app:not_imitation}

A natural concern about distilling from an automatic pipeline is that the probe might learn to reproduce the pipeline's idiosyncrasies rather than anything about uncertainty. Since both training and pipeline-label evaluation share the same source, a high AUROC against pipeline labels alone cannot distinguish the two. Three observations separate them.

\textbf{Human-grounded risk reduction.}
The risk--coverage result (\textbf{Sec.\,\ref{sec:exp:risk_coverage}}) is measured against human judgments that played no part in training. Rejecting spans by predicted uncertainty drives the human-judged error rate monotonically down, from 18.6\% at full coverage to 6.3\% at 40\% coverage. A probe that had memorized pipeline artifacts uncorrelated with real error would produce a flat curve here, because its ranking would carry no information about human-judged correctness.

\textbf{Independent human validation.}
The annotation study (\textbf{App.\,\ref{app:human_study}}) shows the pipeline labels themselves track human judgment ($\kappa = 0.618$; monotone error rate across $u^*$ bins), so the training signal is not arbitrary noise to begin with.

\textbf{Complementary information source.}
The probe's input is the LLM's own hidden states, which encode the model's internal state at generation time. The label pipeline observes only sampled output text and cannot access this. The probe therefore learns to decode an internal signal that is correlated with, but not a copy of, its supervision --- consistent with its ability to rank spans that the pipeline itself scores identically.

\textbf{Cross-domain behaviour.}
A probe trained on the mixed-domain training split retains performance on held-out domain slices (\textbf{App.\,\ref{app:entity_popularity}}), indicating transferable uncertainty structure rather than domain-specific pipeline bias.

We emphasize the limit of this evidence: it is end-to-end. It does not isolate the contribution of each pipeline component, which remains open (see \textbf{Limitations}, Sec.\,\ref{sec:conclusion}).

\subsection{False-Positive Root-Cause Audit}
\label{app:fp_audit}

The human study establishes \emph{that} 22.3\% of high-uncertainty spans are factually correct.
This audit asks \emph{why} the pipeline scored them as uncertain.
The two analyses answer different questions and are complementary: the human study measures downstream utility of the score, whereas the audit is a diagnostic of the pipeline's internal mechanism and does not re-judge factual correctness.

\textbf{Sampling.}
We randomly sampled \textbf{45} of the false-positive spans (those with $u^* \geq 0.5$ but judged \textsc{correct} by the annotator) for manual root-cause audit.
Exhaustive auditing was infeasible, as each case requires reading all 20 sampled responses together with their per-sample verification records.
All percentages below are over this $N = 45$ audited subset, not over the full false-positive set.

\textbf{Procedure.}
For each audited span the auditors had access to the original query; the complete primary response with the span highlighted; all 20 sampled responses; and the per-sample verification record, i.e.\ for every sample the verdict (\textsc{supported} / \textsc{contradicted} / \textsc{not\_mentioned}) together with the evidence string the verifier cited.
They then read the claim text, inspected every \textsc{contradicted} verdict to determine whether the cited evidence genuinely conflicts with the claim or whether the conflict was manufactured by a pipeline artifact, and assigned a single primary root cause.

\begin{table}[h]
  \caption{Root causes of pipeline false positives ($u^* \geq 0.5$, human-judged \textsc{correct}), over a random audit sample of $N = 45$.}
  \label{tab:fp_audit}
  \centering
  \small
  \begin{tabular}{p{0.60\textwidth}rr}
    \toprule
    \textbf{Category} & \textbf{Share} & \textbf{Count} \\
    \midrule
    \texttt{GENUINE\_UNCERTAINTY} --- the sample texts genuinely conflict about the same proposition & 88.9\% & 40 \\
    \texttt{GRANULARITY\_MISMATCH} --- the decomposed claim is narrower or broader than what the samples assert, so verdicts address a different proposition & 4.4\% & 2 \\
    \texttt{PARAPHRASE\_VARIANCE} --- samples express the same fact in different surface forms and the verifier read the wording difference as a contradiction & 4.4\% & 2 \\
    \texttt{CLAIM\_AMBIGUITY} --- the claim admits two valid readings, so samples can disagree without either being wrong & 2.2\% & 1 \\
    \bottomrule
  \end{tabular}
\end{table}

\textbf{Interpretation.}
The large majority (88.9\%) of audited false positives are not pipeline defects.
They are spans where the model's 20 samples substantively disagreed: the uncertainty score was earned by real cross-sample variance, and the claim merely happened to be correct.
This is the behaviour a continuous uncertainty score \emph{should} exhibit --- the model finds the fact unreliable even though it turns out to be true --- and it is also why $u^*$ must not be read as a factuality label (see \textbf{Limitations}, Sec.\,\ref{sec:conclusion}).
Only 11.1\% of the audited cases (5 of 45) are attributable to pipeline mechanisms: granularity mismatch and paraphrase variance in claim decomposition and verification, plus one ambiguous claim.
That fraction is small but concrete, and it defines the improvement target for future versions of \bench{}.
We note the audit is a single-annotator diagnostic over a modest sample; the category shares should be read as indicative rather than precise.

\subsection{Multi-Seed Stability}
\label{app:multiseed}

To quantify sensitivity to training randomness, we retrained the probe for every backbone with four additional random seeds (123, 456, 789, 1024) beyond the default seed 42.
All seeds are evaluated on the \emph{same} per-backbone span set used in \textbf{Tab.\,\ref{tab:span}} (the unified Hungarian-matched set), so the reported variance isolates training randomness and excludes evaluation-set differences. AUROC is stable across seeds (std $\leq 0.017$), confirming that the probe's ability to separate high- from low-uncertainty spans does not depend on a fortunate initialization.
MAE (std $\leq 0.021$) and $\rho_s$ (std $\leq 0.030$) are likewise stable.
The slightly wider spread on Qwen3-4B is consistent with its smaller capacity, but all backbones retain strong ranking performance.

\begin{table}[h]
  \caption{Span-level performance across five random seeds (42, 123, 456, 789, 1024), mean $\pm$ std. \textbf{Tab.\,\ref{tab:span}} reports seed 42, the default used throughout the paper.}
  \label{tab:multiseed}
  \centering
  \small
  \begin{tabular}{lccc}
    \toprule
    \textbf{Backbone} & \textbf{AUROC} & \textbf{MAE} & $\boldsymbol{\rho_s}$ \\
    \midrule
    Qwen3-14B      & 0.930 $\pm$ 0.007 & 0.117 $\pm$ 0.007 & 0.663 $\pm$ 0.013 \\
    Qwen3-8B       & 0.925 $\pm$ 0.005 & 0.137 $\pm$ 0.007 & 0.682 $\pm$ 0.006 \\
    Qwen3-4B       & 0.926 $\pm$ 0.017 & 0.149 $\pm$ 0.021 & 0.721 $\pm$ 0.030 \\
    Qwen3-30B-A3B  & 0.928 $\pm$ 0.010 & 0.114 $\pm$ 0.010 & 0.634 $\pm$ 0.014 \\
    Mistral-7B     & 0.884 $\pm$ 0.014 & 0.146 $\pm$ 0.012 & 0.616 $\pm$ 0.010 \\
    \bottomrule
  \end{tabular}
\end{table}

\subsection{Additional Hidden-State Baselines: HaloScope and HaMI}
\label{app:new_baselines}

We implement two recent hidden-state hallucination detectors as additional baselines.
Both are sequence-level detectors that operate on internal representations, which makes them the closest prior work to \ours{}'s approach.

\textbf{HaloScope}~\citep{du2024haloscope} applies PCA dimensionality reduction followed by logistic regression on mean-pooled last-layer hidden states.
We train it on the full training split (${\sim}$17K sequences).

\textbf{HaMI}~\citep{niu2026robust} uses gated-attention multiple-instance learning over token-level hidden states with a binary cross-entropy objective.
We train it on a 5K random subset of the training split; this is sufficient for the small MIL network and is required because per-sample token-level processing is memory-bound.

\textbf{Adaptation to span-level evaluation.}
Both methods emit a single score per sequence, so we apply the same broadcast protocol used for the other sequence-level baselines in \textbf{Tab.\,\ref{tab:span}} (SelfCheckGPT-NLI, P(True)): the sequence score is assigned to every span in that response.

\textbf{Findings.}
HaloScope outperforms HaMI on most backbones (average AUROC 0.820 vs.\ 0.794), which suggests that mean-pooling yields a more stable sequence representation than MIL attention for a sequence-level classification target.
Both are competitive with SelfCheckGPT-NLI (average 0.804), confirming that hidden-state methods can match multi-sample approaches without additional generation cost.
\ours{} nonetheless leads HaloScope by 11.1 AUROC points on average, and the gap is widest on Spearman correlation (average $\rho_s$ 0.683 vs.\ 0.492).
The reason is structural rather than incidental: broadcasting assigns an identical score to correct and incorrect spans within the same response, so even a perfect sequence-level detector cannot rank spans within a response.
This limitation is what span-native prediction removes.

\subsection{Threshold Selection and Deployment}
\label{app:thresholds}

\textbf{Interpreting the score.}
\ours{}'s output approximates the probability that a span is incorrect, so $0.5$ 
(more likely wrong than right)
is a natural default.
For Qwen3-14B the full model reaches ECE $= 0.013$ on \emph{raw}, uncalibrated scores (\textbf{Tab.\,\ref{tab:ablation}}), i.e.\ it is close to self-calibrated: the Mixture of Beta head needs only $T = 1.05$ of temperature correction, against $T = 0.83$ for a single Beta (\textbf{App.\,\ref{app:mob_analysis}}).
Predicted probabilities therefore track empirical error rates closely enough to compare scores across different inputs from the same backbone.

\textbf{Choosing an operating point.}
The risk--coverage curve (\textbf{Tab.\,\ref{tab:risk_coverage}}) converts an error budget directly into a threshold: accepting at most a 10\% factual error rate corresponds to ${\sim}$80\% coverage, i.e.\ rejecting the top-20\% most uncertain spans.
The monotone relationship between $u^*$ and human-judged error (\textbf{Tab.\,\ref{tab:human_bins}}) guarantees that any threshold produces meaningful separation, with no inversion to worry about: filtering at $u \geq 0.4$ removes predominantly incorrect spans while retaining spans whose error rate is 5.8\%.
As rough starting points by application risk profile: ${\approx}0.3$ for high-stakes settings such as medical or legal review, where flagging more spans for human verification is worth the added cost; ${\approx}0.5$ for customer-facing assistants; and ${\approx}0.7$ for low-stakes uses such as creative writing support, where only near-certain errors should be surfaced.

\textbf{Comparability across generations.}
Within a backbone, calibrated scores are comparable across inputs: 0.6 on one query and 0.6 on another indicate similar uncertainty.
The probe reads the hidden states of whatever text was generated, so it is agnostic to the decoding strategy; our experiments use greedy decoding, and verifying calibration under nucleus or beam search is straightforward future work.
Multi-seed results (\textbf{App.\,\ref{app:multiseed}}) show the calibrated scores are also reproducible across training runs.

\textbf{Deployment recipe.}
(1)~Train a per-backbone probe (${\sim}$3--9 GPU-hours, one-time).
(2)~Fit temperature scaling on a held-out dev split to obtain calibrated probabilities.
(3)~Plot the risk--coverage curve on a representative validation set and read off the threshold matching the application's error budget.
(4)~At inference, run the single LLM forward pass, extract hidden states, apply the probe ($<$3\% added latency), and surface spans above the threshold.

\subsection{Broader Impacts}
\label{app:broader_impacts}

Reliable span-level uncertainty estimation can improve the trustworthiness of LLM-powered systems in high-stakes domains such as healthcare, legal analysis, and education, because it lets a user verify the specific claims a model is unsure about instead of re-checking an entire response or discarding it wholesale. The localization is the point: a sequence-level score forces an all-or-nothing decision, whereas span-level scores support targeted verification.

The same property invites overreliance. A user shown a partially highlighted response may reasonably infer that the unhighlighted remainder has been checked, which it has not. This failure mode is sharpest for the blind spot documented in \textbf{App.\,\ref{app:human_study}}: consistency-based uncertainty is lowest precisely on \emph{confident} hallucinations, where a model repeats the same incorrect claim across all samples. An interface built on \ours{} should therefore communicate that low uncertainty is not a correctness guarantee, and should avoid visual conventions that imply verification of unflagged text.

\ours{} estimates uncertainty from hidden-state patterns and does not certify factual correctness. It should complement, not replace, human judgment and external verification. We release the probe and benchmark to support research on fine-grained uncertainty; the released artifacts contain uncertainty annotations over model-generated text and no personal data.

\subsection{Extending Beyond Factual Uncertainty}
\label{app:extensions}

\ours{}'s architecture is task-agnostic: it consumes hidden states and predicts uncertainty over text spans.
Porting it to a new domain requires only a meaningful notion of ``span'' and a way to obtain span-level labels.
For \emph{mathematical reasoning}, spans could be individual derivation steps (e.g.\ ``Since $x = 3$, we have $2x = 6$''), with labels from step-level verification; the multi-sample construction applies directly, since sampling several reasoning chains reveals where they diverge.
For \emph{code generation}, spans could be statements or blocks, with labels from execution-based verification such as per-segment test outcomes.
We focus on factual generation here because it offers mature automatic verification, because factual hallucination is the most common failure mode in deployed systems, and because it permits rigorous benchmarking across diverse domains.
The architectural components (DETR decoder, MoB uncertainty head, UCIR) are domain-agnostic and would transfer without modification; obtaining reliable step- or execution-level labels is the substantive work.

\subsection{Downstream Applications and Future Directions}
\label{app:future_directions}

\textbf{Uncertainty-aware verification.}
Verifying a long, fact-dense response in full requires many retrieval calls and model invocations, yet not every claim needs the same verifications.
Span-level scores allow a system to send only high-uncertainty spans to search tools or verifier models, so that limited compute goes where errors are most likely.
The risk-tiered thresholds in \textbf{App.\,\ref{app:thresholds}} extend naturally to tiered verification: use low-uncertainty spans directly, apply lightweight checks to intermediate ones, and reserve external retrieval or cross-model checking for the highest-uncertainty spans.
Since low uncertainty does not guarantee correctness (\textbf{App.\,\ref{app:broader_impacts}}), the policy must still be set according to task risk and calibration quality.

\textbf{Uncertainty-guided self-refinement.}
Existing self-refinement methods typically critique and regenerate the entire response.
When only a few spans are unreliable, regenerating everything wastes compute and can corrupt claims that were already correct.
\ours{} makes targeted refinement possible: locate the high-uncertainty spans, then verify, edit, or regenerate only those.
Open questions include when a span's uncertainty justifies refinement, how to keep local edits consistent with the surrounding context, and how to confirm that a refinement actually reduced error risk rather than just changed the text.

\textbf{Uncertainty-aware multi-agent communication.}
In multi-agent systems, one agent's output often becomes another's input, for example passing from a retrieval agent to a reasoning agent to a planning agent.
Today agents exchange natural language only, so a receiving agent has to judge for itself which claims to trust.
If the sending agent also attached a span-level uncertainty score to each claim, the receiver could accept a claim, request supporting evidence, or route it to another agent for cross-checking.
The system as a whole could likewise use these signals to allocate tasks, compute, and reasoning paths.
Realizing this raises problems that \ours{} does not solve.
Our scores are calibrated within a single backbone (\textbf{App.\,\ref{app:thresholds}}), but agents may run on different base models, so ``0.9'' from one probe need not mean the same as ``0.9'' from another.
Cross-model calibration, and preventing miscalibrated signals from propagating and compounding across agents, remain open.
We view span-level uncertainty as a starting point for this line of work, not a solution to it.

\subsection{Benchmark Statistics}
\label{app:benchmark_stats}

\begin{table}[h]
  \caption{Statistics of \bench{} (Qwen3-14B backbone), reported as \textbf{raw (unfiltered)} span counts as emitted by the distillation pipeline. The benchmark spans five domains with stratified splits preserving domain and uncertainty-bin proportions. All labels are soft uncertainty scores from the multi-sample distillation pipeline.}
  \label{tab:benchmark_stats}
  \centering
  \small
  \begin{tabular}{lrrrr}
    \toprule
    \textbf{Split} & \textbf{Prompts} & \textbf{Spans} & \textbf{Avg.\ Spans/P} & \textbf{Avg.\ Words/Resp.} \\
    \midrule
    Train          & 17{,}494 & 272{,}086 & 15.6 & 173 \\
    Dev            & 500      & 7{,}967   & 15.9 & --- \\
    Test           & 2{,}000  & 31{,}332  & 15.7 & 174 \\
    \midrule
    \textbf{Total} & \textbf{19{,}994} & \textbf{311{,}385} & \textbf{15.6} & --- \\
    \bottomrule
  \end{tabular}
\end{table}

\textbf{Raw vs.\ Filtered Span Counts.}
The raw pipeline emits 311{,}385 spans across all splits (table above).
Spans that no sample either supports or contradicts have no defined $u^*$ (\textbf{App.\,\ref{app:distillation}}, Stage 4) and are removed, leaving ${\sim}$293K valid spans in the released benchmark, of which ${\sim}$256K fall in the training split.
All span counts quoted elsewhere in this paper refer to this filtered set unless explicitly marked as raw.

\begin{table}[h]
  \caption{Per-domain breakdown of \bench{}. Biography and long-form QA are the two largest domains; ELI5 provides open-ended diversity; FELM and TriviaQA contribute factoid and reasoning challenges.}
  \label{tab:benchmark_domain}
  \centering
  \small
  \begin{tabular}{lrrrrr}
    \toprule
    \textbf{Domain} & \textbf{Train} & \textbf{Dev} & \textbf{Test} & \textbf{Total} & \textbf{Avg.\ Words/Resp.} \\
    \midrule
    Biography       & 1{,}750  & 50  & 200  & 2{,}000  & 323 \\
    Long-Form QA    & 7{,}348  & 210 & 841  & 8{,}399  & 122 \\
    TriviaQA        & 4{,}375  & 124 & 500  & 4{,}999  & 81  \\
    ELI5            & 2{,}624  & 76  & 299  & 2{,}999  & 349 \\
    FELM            & 1{,}397  & 40  & 160  & 1{,}597  & 217 \\
    \midrule
    \textbf{Total}  & \textbf{17{,}494} & \textbf{500} & \textbf{2{,}000} & \textbf{19{,}994} & \textbf{173} \\
    \bottomrule
  \end{tabular}
\end{table}

\textbf{Domain Characteristics.}
The five domains cover a spectrum of hallucination patterns:
\textit{Biography} (10\%) generates long-form text about entities stratified by Wikipedia popularity (30\% head / 40\% torso / 30\% tail), where tail entities exhibit ${\sim}$3$\times$ higher hallucination rates.
\textit{Long-form QA} (42\%) uses real Google search queries from Natural Questions~\citep{kwiatkowski2019natural}, predominantly factoid (98\%).
\textit{TriviaQA} (25\%) provides trivia questions with verified reference answers~\citep{joshi2017triviaqa}.
\textit{ELI5} (15\%) contains explanatory ``why/how'' questions from Reddit, producing subtle reasoning errors rather than factual fabrications.
\textit{FELM} (8\%) aggregates challenging examples from FELM~\citep{zhao2023felm}, TruthfulQA~\citep{lin2022truthfulqa}, and HaluEval~\citep{li2023halueval}, targeting known misconceptions and math reasoning.

\textbf{Uncertainty Distribution.}
Splits are stratified by domain $\times$ uncertainty bin (low: $u < 0.1$, mid: $0.1 \leq u < 0.4$, high: $u \geq 0.4$) to ensure balanced representation.
Across the full dataset, approximately 62\% of spans have low uncertainty, 20\% mid, and 18\% high.

\textbf{Quality Validation.}
To validate the automatic labels, three co-authors independently reviewed a random sample of 200 test examples (10\% of the test set), checking span boundary correctness, factuality label accuracy, and uncertainty score plausibility.
The reviewers confirmed that the pipeline produces reliable labels, with disagreements primarily limited to temporally sensitive claims and ambiguous quantitative statements (see \textbf{App.\,\ref{app:annotation}} for the full protocol).

\section{Implementation Details}
\label{app:implementation}

\subsection{Loss Function Details}
\label{app:loss}

This section provides the full formulations of each loss component in Eq.\,\ref{eq:total_loss}.

\textbf{Span Regression Loss.}
The regression loss combines L1 distance and Generalized IoU (GIoU)~\citep{rezatofighi2019generalized} over matched span pairs from the Hungarian assignment $\hat{\sigma}$:
\begin{equation}
  \mathcal{L}_{\text{reg}} = \sum_{i=1}^{M} \bigl[ \lVert \hat{\vb}_i - \vb_{\hat{\sigma}(i)}^* \rVert_1 + \lambda_{\text{giou}} \cdot \mathcal{L}_{\text{GIoU}}(\hat{\vb}_i, \vb_{\hat{\sigma}(i)}^*) \bigr],
\end{equation}
where $\hat{\vb}_i = (\hat{b}_i, \hat{e}_i)$ are predicted boundaries, $\vb^*$ are ground-truth boundaries, and $M$ is the number of matched pairs.
GIoU extends standard IoU to handle non-overlapping spans by penalizing the gap between them.

\textbf{Validity Classification Loss.}
Binary cross-entropy over all $N$ span queries:
\begin{equation}
  \mathcal{L}_{\text{val}} = -\frac{1}{N} \sum_{k=1}^{N} \bigl[ y_k \log p_k^{\text{valid}} + (1 - y_k) \log (1 - p_k^{\text{valid}}) \bigr],
\end{equation}
where $y_k = 1$ if span query $k$ is matched to a ground-truth span and $y_k = 0$ otherwise (unused slot).

\textbf{MoB Negative Log-Likelihood.}
See Eq.\,\ref{eq:mob_nll} in the main text.
To avoid numerical instability at $u = 0$ or $u = 1$, we clamp ground-truth labels to $[\epsilon, 1-\epsilon]$ with $\epsilon = 10^{-4}$.

\textbf{Consistency Loss.}
The consistency loss enforces agreement between the span-aggregated sequence-level prediction and the ground-truth sequence-level uncertainty:
\begin{equation}
  \mathcal{L}_{\text{con}} = \lVert \hat{u}_{\text{seq}} - u_{\text{seq}}^* \rVert_2^2,
\end{equation}
where $u_{\text{seq}}^* = \frac{1}{K} \sum_{k=1}^{K} u_k^*$ is the mean ground-truth span uncertainty.
Gradients from $\mathcal{L}_{\text{con}}$ are detached from the span-level predictions to prevent the sequence-level objective from interfering with span-level learning.

\textbf{Contrastive Ranking Loss.}
See Eq.\,\ref{eq:rank} in the main text.
We use margin $m = 0.1$.
In practice, rather than enumerating all valid pairs, we employ \textit{stratified sampling}: spans are partitioned into a high-uncertainty group ($u^* > \tau_{\text{hi}} = 0.3$) and a low-uncertainty group ($u^* < \tau_{\text{lo}} = 0.1$), and up to 256 (high, low) pairs are randomly sampled per batch.
If either group contains fewer than 2 spans, we fall back to the top-25\% vs.\ bottom-25\% of spans ranked by $u^*$.
This strategy focuses the ranking signal on clearly separable pairs and avoids noisy gradients from near-identical labels.

\textbf{Loss Weights.}
The default loss weights are: $\lambda_{\text{reg}} = 5.0$, $\lambda_{\text{val}} = 2.0$, $\lambda_{\text{uq}} = 4.0$, $\lambda_{\text{con}} = 1.0$, $\lambda_{\text{rank}} = 0.5$, $\lambda_{\text{aux}} = 0.4$.
During the warmup phase (epochs 1--15), only $\mathcal{L}_{\text{reg}}$ and $\mathcal{L}_{\text{val}}$ are active.

\subsection{Model Specifications}
\label{app:model_specs}

\textbf{Tab.\,\ref{tab:app_models}} summarizes the LLM backbones used in our experiments, and \textbf{Tab.\,\ref{tab:app_probe_specs}} details the \ours{} probe configuration selected for each model via hyperparameter sweep (\textbf{Sec.\,\ref{app:hp_sensitivity}}).

\begin{table}[h]
  \centering
  \small
  \begin{tabular}{lccccccc}
    \toprule
    \textbf{Model} & \textbf{Type} & \textbf{Params} & \textbf{Layers} & \textbf{$d_\text{model}$} & \textbf{Heads} & \textbf{KV Heads} & \textbf{Vocab} \\
    \midrule
    Qwen3-14B       & Dense & 14.8B & 40 & 5120 & 40 & 8 & 151,936 \\
    Qwen3-8B        & Dense & 8.2B  & 36 & 4096 & 32 & 8 & 151,936 \\
    Qwen3-4B        & Dense & 4.0B  & 36 & 2560 & 32 & 8 & 151,936 \\
    Qwen3-30B-A3B   & MoE   & 30.5B (3.3B active) & 48 & 2048 & 16 & 4 & 151,936 \\
    Mistral-7B-Instruct-v0.3 & Dense & 7.2B  & 32 & 4096 & 32 & 8 & 32,768 \\
    \bottomrule
  \end{tabular}
  \caption{LLM backbone specifications. All models use grouped-query attention (GQA). Qwen3-30B-A3B is a Mixture-of-Experts model with 3.3B active parameters per token.}
  \label{tab:app_models}
\end{table}

\begin{table}[h]
  \centering
  \small
  \begin{tabular}{lcccccc}
    \toprule
    \textbf{Model} & \textbf{Probe Params} & \textbf{Layers Used} & \textbf{$d_\text{proj}$} & \textbf{$n_q$} & \textbf{Batch} & \textbf{LR} \\
    \midrule
    Qwen3-14B      & 27.04M & [22, 24, 26] & 512 & 32 & 16 & $10^{-4}$ \\
    Qwen3-8B       & 26.51M & [18, 20, 22] & 512 & 32 & 16 & $10^{-4}$ \\
    Qwen3-4B       & 17.18M & [22, 24, 26] & 512 & 32 & 16 & $10^{-4}$ \\
    Qwen3-30B-A3B  & 25.46M & [30, 34, 38] & 512 & 32 & 16 & $10^{-4}$ \\
    Mistral-7B     & 26.50M & [12, 14, 16] & 512 & 16 & 32 & $2{\times}10^{-4}$ \\
    \bottomrule
  \end{tabular}
  \caption{\ours{} probe configurations (best per model from hyperparameter sweep). All probes share: 2-layer encoder, 3-layer decoder, 8 attention heads, $d_\text{ffn}{=}2048$, MoB $K{=}3$, UCIR $\alpha{=}0.7$, span enrichment with gate fusion.}
  \label{tab:app_probe_specs}
\end{table}

\subsection{Hyperparameter Sensitivity}
\label{app:hp_sensitivity}

We conduct a systematic hyperparameter sweep with 24--28 configurations per model, varying layer selection, projection dimension ($d_{\text{proj}} \in \{256, 512\}$), number of span queries ($n_q \in \{16, 32\}$), batch size ($\in \{16, 32\}$), and learning rate ($\in \{5{\times}10^{-5}, 10^{-4}, 2{\times}10^{-4}\}$).
\textbf{Tab.\,\ref{tab:hp_sensitivity}} summarizes the sensitivity of dev AUROC to each hyperparameter, averaged across all five models.

\begin{table}[h]
  \caption{Hyperparameter sensitivity analysis. For each hyperparameter, we report the mean dev AUROC across all configurations using that value, averaged over five models. $\Delta$ denotes the range (max $-$ min) across values.}
  \label{tab:hp_sensitivity}
  \centering
  \small
  \begin{tabular}{llcc}
    \toprule
    \textbf{Hyperparameter} & \textbf{Values} & \textbf{Mean AUROC} & $\boldsymbol{\Delta}$ \\
    \midrule
    $d_{\text{proj}}$ & 256 / 512 & 0.898 / 0.916 & 0.018 \\
    $n_q$ (span queries) & 16 / 32 & 0.905 / 0.910 & 0.005 \\
    Batch size & 16 / 32 & 0.908 / 0.907 & 0.001 \\
    Learning rate & $5{\times}10^{-5}$ / $10^{-4}$ / $2{\times}10^{-4}$ & 0.904 / 0.910 / 0.908 & 0.006 \\
    \bottomrule
  \end{tabular}
\end{table}

\textbf{Key Findings.}
(1)~\textbf{Projection dimension} is the most impactful hyperparameter ($\Delta = 0.018$): $d_{\text{proj}} = 512$ consistently outperforms 256 across all models, suggesting that the uncertainty signal requires sufficient representational capacity.
(2)~\textbf{Batch size} has negligible effect ($\Delta = 0.001$), indicating that the training dynamics are stable across batch sizes.
(3)~\textbf{Learning rate} $10^{-4}$ is optimal for most models, though Mistral-7B benefits from a higher rate ($2{\times}10^{-4}$), likely reflecting differences in hidden-state scale across model families.

\subsection{Training Curves}
\label{app:training_curves}

\textbf{Tab.\,\ref{tab:app_training_curves}} reports the training dynamics for the best \ours{} probe configuration on each LLM backbone.
All models use a two-phase schedule: 15 warmup epochs (span detection only) followed by up to 25 joint epochs (span detection + uncertainty estimation), with early stopping (patience~5) on dev AUROC.

\begin{table}[h]
  \centering
  \small
  \begin{tabular}{lcccccc}
    \toprule
    \textbf{Model} & \textbf{Epochs} & \textbf{Best Ep.} & \textbf{Dev AUROC} & \textbf{Dev MAE} & \textbf{Dev $\rho_s$} & \textbf{Wall Time} \\
    \midrule
    Qwen3-14B      & 38 (15w+23j) & 33 & 0.9412 & 0.1031 & 0.7932 & 9.3h \\
    Qwen3-8B       & 40 (15w+25j) & 39 & 0.9275 & 0.1257 & 0.7699 & 5.2h \\
    Qwen3-4B       & 38 (15w+23j) & 33 & 0.9444 & 0.1219 & 0.8225 & 4.0h \\
    Qwen3-30B-A3B  & 39 (15w+24j) & 34 & 0.9350 & 0.1020 & 0.7433 & 3.7h \\
    Mistral-7B     & 28 (15w+13j) & 23 & 0.9279 & 0.1142 & 0.7312 & 2.7h \\
    \bottomrule
  \end{tabular}
  \caption{Training summary for the best \ours{} probe per model. ``w'' = warmup epochs, ``j'' = joint epochs. All training on a single NVIDIA H100 80GB GPU. Dev metrics are reported at the best epoch (selected by AUROC).}
  \label{tab:app_training_curves}
\end{table}

\textbf{Key observations.}
(1)~All models exhibit a sharp phase transition at the warmup$\to$joint boundary (epoch~16): dev AUROC jumps from ${\sim}0.55$--$0.65$ to ${\sim}0.85$+ within 2--3 joint epochs, confirming that the warmup phase successfully pre-trains span detection before uncertainty estimation begins.
(2)~Qwen3-4B achieves the highest dev AUROC (0.9444) despite having the smallest probe (17.18M parameters), suggesting that smaller LLMs with lower $d_\text{model}$ can produce equally informative hidden-state representations for uncertainty estimation.
(4)~Mistral-7B trains fastest (2.7h) due to its larger batch size (32) and early convergence (epoch~23).

\subsection{Computational Cost}
\label{app:cost}

\textbf{Tab.\,\ref{tab:cost}} provides a detailed computational cost breakdown for \ours{}.

\begin{table}[h]
  \caption{Computational cost breakdown. Data construction is a one-time cost shared across all models. Probe training and inference are per-model costs. All timings on NVIDIA H100 80GB GPUs.}
  \label{tab:cost}
  \centering
  \small
  \begin{tabular}{llr}
    \toprule
    \textbf{Stage} & \textbf{Details} & \textbf{Cost} \\
    \midrule
    \multicolumn{3}{l}{\emph{One-time data construction (per LLM backbone)}} \\
    Multi-sample generation & 1 greedy + 20 samples $\times$ 20K prompts, 8$\times$H100 & ${\sim}$40 GPU-h \\
    Claim decomposition + verification & ${\sim}$700K API calls (Claude Opus 4.6) & ${\sim}$\$1{,}000 \\
    Hidden state extraction & Single forward pass, 20K prompts & ${\sim}$0.5 GPU-h \\
    \midrule
    \multicolumn{3}{l}{\emph{Probe training (single GPU)}} \\
    Qwen3-14B  & 38 epochs (15w + 23j), 14 min/epoch & 9.3h \\
    Qwen3-8B   & 40 epochs (15w + 25j), 8 min/epoch  & 5.2h \\
    Qwen3-4B   & 38 epochs (15w + 23j), 6 min/epoch  & 4.0h \\
    Qwen3-30B-A3B & 39 epochs (15w + 24j), 6 min/epoch & 3.7h \\
    Mistral-7B    & 28 epochs (15w + 13j), 6 min/epoch & 2.7h \\
    \midrule
    \multicolumn{3}{l}{\emph{Inference (per prompt)}} \\
    LLM forward pass & Response generation (greedy) & 2--5s \\
    Hidden state caching & Extract from 3 layers & $<$10ms \\
    \ours{} probe & Projection + DETR + UCIR & $<$50ms \\
    \textbf{Total overhead} & \textbf{Probe / LLM forward pass} & \textbf{$<$3\%} \\
    \bottomrule
  \end{tabular}
\end{table}

\textbf{Training Efficiency.}
The \ours{} probe trains in 2.7--9.3 hours on a single H100 GPU, depending on the backbone model size (which determines $d_{\text{model}}$ and thus the input dimension).
The 14B model requires the longest training due to its larger projection layer ($5120 \to 512$), while Mistral-7B trains fastest due to its larger batch size (32) and early convergence.
The LLM backbone is \emph{frozen} during training: only the lightweight probe (17--27M parameters) is updated, requiring no gradient computation through the LLM.

\textbf{Inference Overhead.}
At inference time, \ours{} adds $<$3\% latency to the LLM forward pass.
The dominant cost is the LLM generation itself (2--5s per prompt); the probe computation (projection, DETR encoding/decoding, UCIR refinement) takes $<$50ms.
This is $10$--$20{\times}$ faster in total wall-clock time than sampling-based methods (e.g., Semantic Entropy, SelfCheckGPT) that require 10--20 additional forward passes.

\textbf{Comparison with Baselines.}
Semantic Entropy requires $S = 10$ sampled responses plus NLI-based clustering, costing ${\sim}10{\times}$ the generation budget.
SelfCheckGPT-NLI similarly requires $S = 10$ samples plus $S \times K$ NLI inference calls (where $K$ is the number of sentences).
FActScore requires claim decomposition and verification via an external LLM, adding ${\sim}$5s per claim.
In contrast, \ours{} operates on cached hidden states from a single forward pass, making it the most efficient method among all approaches that achieve AUROC $> 0.85$.

\end{document}